\documentclass[preprint,12pt,preprint]{elsarticle}

\usepackage{graphicx}%
\usepackage{multirow}%
\usepackage{amsmath,amssymb,amsfonts}%
\usepackage{amsthm}%
\usepackage{amsmath}
\usepackage{mathrsfs}%
\usepackage[title]{appendix}%
\usepackage{xcolor}%
\usepackage{textcomp}%
\usepackage{manyfoot}%
\usepackage{booktabs}%
\usepackage{algorithm}%
\usepackage{algorithmicx}%
\usepackage{algpseudocode}%
\usepackage{listings}%

\usepackage{url}
\usepackage[capitalise]{cleveref}
\usepackage{multirow}
\usepackage{adjustbox}
\usepackage{todonotes}
\usepackage{placeins}
\newcommand\norm[1]{\left\lVert#1\right\rVert}
\creflabelformat{equation}{#2\textup{#1}#3}

\usepackage{lineno}

\renewcommand{\cite}{\citep}

\journal{Applied Soft Computing}

\begin{document}

\begin{frontmatter}



\title{CINDI: Conditional Imputation and Noisy Data Integrity with Flows in Power Grid Data} 


\author{David Baumgartner\corref{cor1}}
\ead{david.baumgartner@ntnu.no}

\author{Helge Langseth}

\author{Heri Ramampiaro}

\cortext[cor1]{Corresponding author}

\address{Department of Computer Science, Norwegian University of Science and Technology, Trondheim, Norway}

\begin{abstract}
Real-world multivariate time series, particularly in critical infrastructure such as electrical power grids, are often corrupted by noise and anomalies that degrade the performance of downstream tasks. 
Standard data cleaning approaches often rely on disjoint strategies, which involve detecting errors with one model and imputing them with another.
Such approaches can fail to capture the full joint distribution of the data and ignore prediction uncertainty. 
This work introduces Conditional Imputation and Noisy Data Integrity (CINDI), an unsupervised probabilistic framework designed to restore data integrity in complex time series. 
Unlike fragmented approaches, CINDI unifies anomaly detection and imputation into a single end-to-end system built on conditional normalizing flows. 
By modeling the exact conditional likelihood of the data, the framework identifies low-probability segments and iteratively samples statistically consistent replacements. 
This allows CINDI to efficiently reuse learned information while preserving the underlying physical and statistical properties of the system. 
We evaluate the framework using real-world grid loss data from a Norwegian power distribution operator, though the methodology is designed to generalize to any multivariate time series domain.
The results demonstrate that CINDI yields robust performance compared to competitive baselines, offering a scalable solution for maintaining reliability in noisy environments.
\end{abstract}



\begin{keyword}



Conditional Normalizing Flows \sep Probabilistic Imputation \sep Multivariate Time Series \sep Anomaly Detection \sep Smart Grids

\end{keyword}

\end{frontmatter}



\newcommand{\XFull}{\mathbf{X}}
\newcommand{\xPDF}{p_{\boldsymbol{X}}}
\newcommand{\xBold}{\boldsymbol{x}}
\newcommand{\XFulltrain}{\XFull^{(\text{train})}}
\newcommand{\XFullval}{\XFull^{(\text{val})}}
\newcommand{\XFulleval}{\XFull^{(\text{eval})}}
\newcommand{\XFulltest}{\XFull^{(\text{test})}}

\newcommand{\ZFull}{\mathbf{Z}}
\newcommand{\zPDF}{p_{\boldsymbol{Z}}}
\newcommand{\zBold}{\boldsymbol{z}}

\newcommand{\WFull}{\mathbf{W}}
\newcommand{\wBold}{\boldsymbol{w}}

\newcommand{\YFull}{\mathbf{Y}}
\newcommand{\yBold}{\boldsymbol{y}}

\newcommand{\lFull}{\mathbf{l}}
\newcommand{\lFulltrain}{\mathbf{l}^{(\text{train})}}
\newcommand{\lFulleval}{\mathbf{l}^{(\text{eval})}}
\newcommand{\lFulltest}{\mathbf{l}^{(\text{test})}}

\newcommand{\lBold}{\boldsymbol{l}}

\newcommand{\aBold}{\boldsymbol{a}}
\newcommand{\bBold}{\boldsymbol{b}}
\newcommand{\cBold}{\boldsymbol{c}}
\newcommand{\dBold}{\boldsymbol{d}}
\newcommand{\hBold}{\boldsymbol{h}}
\newcommand{\sBold}{\boldsymbol{s}}
\newcommand{\tBold}{\boldsymbol{t}}

\newcommand{\xd}{\xBold^{1:d}}
\newcommand{\xD}{\xBold^{d+1:D}}
\newcommand{\zd}{\zBold^{1:d}}
\newcommand{\zD}{\zBold^{d+1:D}}

\newcommand{\xtd}{\xBold_t^{1:d}}
\newcommand{\xtD}{\xBold_t^{d+1:D}}
\newcommand{\ztd}{\zBold_t^{1:d}}
\newcommand{\ztD}{\zBold_t^{d+1:D}}

\section{Introduction}
\label{sec:introduction}

Accurately forecasting key values in modern electrical power grids, such as grid loss, is a growing challenge with direct financial implications.
In markets like the Nord Pool exchange, power distribution operators must report precise loss estimates to determine pricing and manage risk.\footnote{See \url{www.nordpoolgroup.com}}
Grid losses are well understood and can be calculated with precision when the grid's configuration is known. 
However, the grid configuration is often unclear or evolving, making it a difficult task to compute losses. 
Further complications exist from stochasticity, the increasing share of renewable energy, changes in consumer demand, and inherent issues with data quality\,\cite{statnett_way_2018,grotmol_robust_2023}.
Consequently, power distribution operators increasingly rely on machine learning models to predict these values.
However, the reliability of these predictive models is strictly bound by the quality of their training data.

This creates a significant bottleneck, as real-world time series data is frequently corrupted by sensor malfunctions, transmission errors, and noise.
These result from equipment malfunctions or data processing issues\,\cite{dalal_day-ahead_2021,wang_deep_2025}.
Crucially, this data should not be cleaned using rudimentary, interpolation-based approaches.
To be useful for downstream tasks, any cleaning process should preserve the underlying physical and statistical properties hidden in the data. 
This must be achieved while correcting corrupted observations.
Standard approaches often treat anomaly detection and data imputation as separate, disjoint tasks\,\cite{kim_probabilistic_2023,wang_task-oriented_2024}.
These methods typically rely on distinct models, limiting their ability to capture the full joint distribution of the data.
This fragmentation limits the model's ability to distinguish between genuine system anomalies and mere data errors\,\cite{jiang_softpatch_2022,wu_understanding_2022,dalal_day-ahead_2021}.

\begin{figure}[ht]
    \centering
    \includegraphics[width=\linewidth]{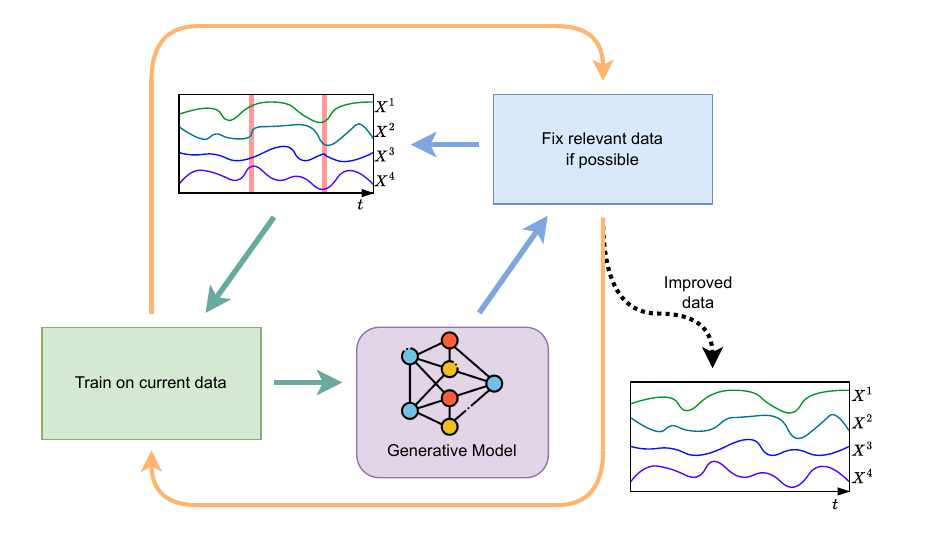}
    \caption{Overview of our CINDI framework based on normalizing flows. 
    The framework alternates between the two states (green and blue), indicated by the orange path, of training and data improvement until convergence.
    At this point, no further changes are made, and an improved dataset is available, as indicated by the black dotted path.
    In the green state, CINDI uses the current dataset and outputs a normalizing flow model. 
    The following blue state uses this model to identify data points that deviate from the expected behavior and then corrects them by generating plausible replacements.
    This process leads to convergence away from detecting unexpected behavior, resulting in improved data for another task.}
    \label{fig:flow-imputation-overview}
\end{figure}

To address this, we propose Conditional Imputation and Noisy Data Integrity (CINDI), a unified framework that integrates detection, correction, and training into a single end-to-end system.
Unlike methods that rely on fragmented architectures, CINDI utilizes a single conditional normalizing flow to model the data distribution of expected grid behavior\,\cite{papamakarios_normalizing_2021,kobyzev_normalizing_2021}.
By learning the exact likelihoods of the data, CINDI can identify low-probability segments (anomalies) and generate plausible, statistically consistent replacements (imputation) in an iterative process.
\cref{fig:flow-imputation-overview} provides a high-level overview of our iterative approach. 

By treating the tasks of detection and correction as intrinsic parts of the probability modeling process, rather than external preprocessing steps, CINDI reuses learned information efficiently\,\cite{shukla_multi-time_2020,xiao_unsupervised_2024,luo_egan_2019}. 
This yields a cleaner, more stable dataset that retains the statistical integrity required for high-stakes forecasting. 
Our experiments demonstrate that this unified, flow-based approach provides a competitive and robust solution for maintaining data integrity in complex, multivariate time series.
In summary, our main contributions are as follows:
\begin{itemize}
    \item CINDI, a novel end-to-end probabilistic framework that detects and removes dataset errors by modeling temporal dependencies in multivariate time series data, utilizing a single conditional normalizing flow model for different tasks.
    \item We apply CINDI to a real-world dataset from a Norwegian power distribution operator to demonstrate its practical applicability, and it is not limited to this dataset.
    \item We conduct extensive experiments and compare our results with those of standard methods, including recent model-based approaches. 
\end{itemize}

The remainder of the paper is organized as follows:
\cref{sec:related-work} reviews related work, followed by \cref{sec:CINDI-framework}, which details the proposed framework and the necessary background. 
Next, \cref{sec:dataset} describes the datasets, while \cref{sec:results} presents the experimental setup and results. 
Finally, \cref{sec:limitations} reflects on the findings, and \cref{sec:conclusion} concludes the paper.

\section{Related Work}
\label{sec:related-work}

\subsection{Learning from noisy data}

Our work is related to learning from data with noise (we use noise equal to error and anomalies in this context), which is an issue in any machine learning task~\cite{hendrycks_using_2018,diakonikolas_distribution-independent_2019,zhang_distilling_2020}. 
These works consider supervised learning and assume relatively balanced label classes. 
This differs significantly from anomaly detection, where errors are rare occurrences that create heavily unbalanced datasets.
Several studies~\cite{han_co-teaching_2018,shen_learning_2019,li_gradient_2020} show that noisy labels give rise to hard-to-learn examples and seek to improve classification performance by explicitly constraining the model on such difficult data.
These approaches need to be tested and verified for anomaly detection, as anomalies are rare occurrences, which gives them less weight in the overall contribution to a model, but requires labeled data.

\subsection{Multivariate Time Series Imputation}

Motivated by the intuition that imputing time series adds valuable knowledge to a dataset, recent works have investigated using diffusion models, transformers, and attention mechanisms\,\cite{wang_deep_2025}. 
\citet{xiao_boundary-enhanced_2025} and \citet{chen_rethinking_2024} both follow a diffusion approach. 
The former utilizes a multi-scale temporal state-space layer, while the latter restricts imputation diversity through negative entropy regularization.
Extensions to transformer models are presented by \citet{gudla_multi-sensor_2026}, \citet{wang_enhanced_2025}, and \citet{liu_timecheat_2025}. 
Common strategies include fusing transformers with different local and global pathways, incorporating spectral information, and adding graph neural networks to model interconnections.
Attention-based methods from \citet{zhang_real-time_2025}, \citet{oh_sting_2021}, and \citet{islam_self-attention-based_2024} incorporate contrastive properties, utilizing bi-directional generative adversarial networks, or combining various attention types.
All of these unite a technical aspect in that they introduce different add-ons or regularization to the underlying base method. 

\citet{hemmer_true_2025} builds on pre-training a mixture-of-experts model with a novel multivariate almost-linear RNN architecture for dynamical system reconstruction, yielding faithful forecasts and outperforming existing foundation models, such as Chronos~\cite{ansari_chronos_2024}, in zero-shot performance.
\citet{seifner_zero-shot_2025} propose, on the other hand, a mathematical approach using ordinary differential equations to construct functions for imputing data, showing promising zero-shot performance without fine-tuning.
These pre-trained models can possess a solid abstraction from well-curated data, providing stability, especially as error levels increase in the data.

\section{CINDI: Conditional Imputation and Noisy Data Integrity}
\label{sec:CINDI-framework}
 
Building on the foundation presented in the introduction (\cref{sec:introduction}), we propose CINDI, a probabilistic end-to-end framework designed to address the challenges discussed in that section by leveraging conditional normalizing flows. 
CINDI systematically detects unusual or corrupted measurements and generates plausible replacements. 
Most prominently, we use one model per iteration for different tasks within our framework. 
This provides a good foundation for ensuring reliable learning and reusing learned expected behavior to improve a dataset.
This section outlines the system’s core principles, emphasizing its adaptation to multivariate time series and its potential to enhance datasets, thereby improving downstream tasks such as anomaly detection and analysis.

\subsection{Background}
\label{subsec:cindi-background}

CINDI uses normalizing flows\,\cite{papamakarios_normalizing_2021}, a generative probabilistic model. 
Normalizing flows are density estimators that learn a data distribution by learning a sequence of transformations that map the data to a known distribution. 
Normalizing flows can both calculate the log-likelihood ($\log p(\xBold)$, normalizing) and sample ($\xBold \sim p$, generate) efficiently. 
They are bijective transformers and require a tractable Jacobian determinant per transformation. 
We denote the normalizing and likelihood of a data point $\xBold$ as $\zPDF(\zBold = F(\xBold))$ and the generation of a data point $\xBold = F^{-1}(\zBold)$.

With multivariate time series data, we utilize conditioned normalizing flows\,\cite{rasul_multivariate_2020}, based on RealNVP\,\cite{dinh_density_2017}, which enables us to capture sequential behavior.
We use RealNVP due to its efficiency in both operation modes and its extensibility.
The inputs to the conditioned normalizing flow are the current observation and a temporal context as defined below.

We refer to a multivariate time series as a sequence $\{\xBold_t\}_{t=1}^T$, where each $\xBold_t \in \mathbb{R}^d$ is a $d$-dimensional vector of variables observed at time $t$, with $d \ge 2$. 
Associated with the time series is a vector $\lFull = (\lBold_1, \lBold_2, \ldots, \lBold_T)$, which is a sequence of binary values of the same length. 
For each $t \in \{1, \ldots, T\}$, $\lBold_t \in \{0, 1\}$ indicates whether the observation at time $t$ is a possible error or expected behavior; specifically, $\lBold_t = 1$ denotes a possible error (anomalous event), while $\lBold_t = 0$ indicates normal behavior.
Given the multivariate time series, we define a windowed sequence with window length $k$ (where $k$ is a hyperparameter) as $\XFull = \left\{ \left(\xBold_t, \wBold_t\right) \right\}_{t=k}^T = \left\{ \left(\xBold_t, \xBold_{t-1}, \ldots, \xBold_{t-k}\right) \right\}_{t=k}^T$,
where each tuple contains the current observation at time $t$ and the $k$ preceding observations as temporal context. 
We adopt the following short notation of the current observation as $\xBold_t$ and the temporal context as $\wBold_t$.
The inputs to the conditioned normalizing flow are the current observation and the temporal context, serving as the conditioning.

We train the normalizing flow unsupervised using negative log-likelihood optimization via the loss function $\mathcal{L}(\XFull) = -\frac{1}{T} \sum_{t=1}^{T} \log(\xPDF(\xBold_t|\wBold_t))$, where $\xBold_t$ depends on $\wBold_t$. 
Since $\xPDF(\cdot)$ is an unknown data distribution, we resort to a normal distribution as the base distribution $\zPDF(\cdot)$ with  $\mu = 0$ and $\Sigma = I$, as the normalization target distribution for the normalizing flow.
The full loss function with the change-of-variable formula is written as
\begin{equation}
    \label{eq:tcNF-objective}
     \mathcal{L}(\XFull) = - \frac{1}{T} \sum_{t=1}^{T} \left[ \log \zPDF(F_\theta(\xBold_t|\wBold_t)) 
    + 
    \sum_{i=1}^{N} \log \vert \det \mathbf{J}(F_{\theta,i})(\xBold_{t,i}|\wBold_t) \vert \right],
\end{equation}
where $NF_\theta$ is a conditional normalizing flow with parameters $\theta$ and temporal context as additional condition $\wBold_t$.
$N$ denotes the number of transformation layers with $F_{\theta,i}$ being the $i$th transformation layer. 
The Jacobian of one such layer is with respect to the input $\xBold_{t,i}$ and the temporal context $\wBold_t$.

CINDI is independent of downstream tasks and focuses on improving data quality. 
The improved data can then be further utilized for different downstream tasks with or without reapplying parts of CINDI. 
The following sections introduce the various building blocks and their inner workings that comprise CINDI.

\subsection{CINDI: Detection}
\label{subsec:cindi-detection}

Given that CINDI is built around conditional normalizing flows, we utilize its probabilistic capabilities to calculate the likelihoods of any data point. 
Therefore, we compare the negative log-likelihood of data points to an average obtained from known expected data points. 
If the negative log-likelihood is significantly higher than expected, then these data points get flagged. 
Hence, we define the threshold $\tau$ as the average likelihood plus twice the standard deviation of $J$ expected data points with $\XFull^{(J,0)} := \{(\xBold_t, \wBold_t) : \lBold_t = 0 \in \mathbf{l}\} \subset \XFull$.
At test time, a data point or a collection of data points gets flagged as detected and therefore marked for imputation if the expectation or the expected average over multiple data points exceeds the threshold $\tau > \mathbb{E}(\xBold_t|\wBold_t)$.

\subsection{CINDI: Imputation}
\label{subsec:cindi-imputation}

CINDI uses the generative functionality of the conditional normalizing flow at its core to generate plausible replacements.
It enables us to generate new data points by sampling from the base distribution or by selecting a point in the latent space and applying the inverse conditional normalizing flow, referred to as $F^{-1}$. 

We obtain data point by $\hat{\xBold}_t = F^{-1}(\zBold_t, \wBold_t)$, with $\zBold_t$ being the center of the base distribution $\zPDF(\zBold_t = \mu, \Sigma)$.
To impute a sequence of data points, the generation process is the same as for a single data point, but the temporal context $\wBold_t$ needs to be updated. 
The first temporal context is sourced from the original data and should contain only expected behavior. 
Every subsequent imputation step requires updating the temporal context with the current generated data point $\hat{\xBold}_t$, as $\wBold_{t+1} = (\hat{\xBold}_t, \xBold_{t-1}, \ldots, \xBold_{t-k+1})$ for the next step. 
This results in a self-regressive chain and the generation of an alternative sub-sequence for a flagged section.
We sample from $\zBold_t = \mu$, because it is the most likely point in the distribution, and with the temporal context, it should produce the most likely expected observation in the data space.

\subsection{CINDI: Model Selection}
\label{subsec:cindi-model-selection}

The model selection process in CINDI is responsible for training multiple models and selecting the one that best fits the given requirements. 
To automatically find a suitable model for a training set $\XFulltrain$, we use CMA-ES\,\cite{auger_restart_2005}, an evolutionary algorithm, to search the hyperparameter space to find a fitting candidate.
Based on the available data, the model selection process needs a function to rank all the candidates.
We provide two functions tailored to the datasets in \cref{sec:dataset}.
The primary objective function is $\phi$, see \cref{eq:aneo-cma-es-objective}, and is used for the real-world dataset.
It requires an evaluation set $\XFulleval = \{(\xBold_u, \wBold_u)\}_{u}^U$ with labels $\lFulleval = \{\lBold_u \in \{0, 1\}\}_{u}^U$, which is independent or non-overlapping with the training data. 
The objective function evaluates a candidate on a balance between detecting the labeled areas via the AUC-ROC (AUC) and VUS-ROC\,\cite{paparrizos_volume_2022} (VUS) metrics, and in being able to reconstruct expected behavior where $\lBold_t = 0 \in \lFulleval$. 
We define it as
\begin{equation}
    \label{eq:aneo-cma-es-objective}
\begin{split}
    \phi(\XFulleval, \lFulleval, M, S) & = 0.3 \cdot (1 - \text{AUC}(\lFulleval, F_\theta(\XFulleval))) \\
    & + 0.7 \cdot (1 - \text{VUS}(\lFulleval, F_\theta(\XFulleval))) \\
    & + \Delta(\XFull^{(\text{eval},0)}, M, S),
\end{split}
\end{equation}
where the model performance is based on the evaluation set and is lower bounded by $0$, which would imply a perfect detection and reconstruction. 
The VUS score receives more weight in the scoring because of its strong capability in scoring range detections compared to point detection in AUC. 
We refer to clean data for the reconstruction metric as $\XFull^{(\text{eval},0)} := \{(\xBold_t, \wBold_t) : \lBold_t = 0 \in \lFulleval\} \subset \XFulleval$ and define the reconstruction metric as
\begin{equation}
    \label{eq:reconstruction-metric}
    \Delta(\XFull^{(\text{eval},0)},M,S) = \frac{1}{|M| \cdot S}\sum^{M}_{m}\sum^{S-1}_{s=0} \norm{\xBold_{m+s} - F^{-1}_\theta(\zBold_{m+s}|\wBold_{m+s})}^2,
\end{equation}
where we calculate the mean reconstruction error from $M := \{u \in U\}$ subsets of length $S$.
The set $M$ contains starting points and remains fixed for all candidates to ensure comparability. 
Next, the normalizing flow is run in inverse and generates new data points $\hat{\xBold}_{m+s}$ from $\zBold_{m+s} = \mu$ starting with $\wBold_{m+0}$ from $\XFull^{(\text{eval},0)}$ as the initial temporal context. 
We update this temporal context $\wBold_{m+s}$ for each $s>0$ continuously with the previous predictions $\hat{\xBold}_{m+s}$, as in \cref{subsec:cindi-imputation}, and create a self-regressive chain. 
This allows us to calculate a reconstruction metric that is lower bounded by $0$ and reflects the model's understanding of expected behavior in a noisy environment, confirming that it remains stable over a period of steps.

The second objective function $\psi$, see \cref{eq:fsb-cma-es-objective}, is for cases when the evaluation set has no associated labels.
The objective function uses, therefore, two negative log-likelihood scores and the reconstruction metric in \cref{eq:reconstruction-metric} and is defined as
\begin{equation}
    \label{eq:fsb-cma-es-objective}
\begin{split}
    \psi(\XFullval, \XFulleval, M, S) & = \lambda \cdot \frac{1}{|\XFullval|} \zPDF(NF_\theta(\XFullval)) \\
    & + \beta \cdot \frac{1}{|\XFulleval|} \zPDF(F_\theta(\XFulleval)) \\
    & + \Delta(\XFulleval, M, S),
\end{split}
\end{equation}
where $\XFullval$ is a hold-out validation set from the training data $\XFulltrain$ for early-stopping the training optimization and to prevent overfitting.
This objective function calculates the average negative log-likelihoods from the validation and evaluation set, and weights them by $\lambda$ and $\beta$, and includes the reconstruction score.
The weighting factors in \cref{eq:fsb-cma-es-objective} are $\lambda = 0.1$ and $\beta = 0.5$ for sequences with $D \leq 10$.
This needs adjustment for sequences with higher dimensions and sequence lengths to weight and balance the different factors accordingly.

\section{Dataset}
\label{sec:dataset}

\subsection{Grid Loss Data}
\label{subsec:aneo-grid-loss-data}

This work addresses the challenge of accurately predicting day-ahead grid loss, a critical task for power distribution operators. 
One important obstacle to reliable prediction is sometimes poor data quality. 
Our goal is to demonstrate, using CINDI, that improving the input data can enhance the performance of downstream tasks. 
To achieve this, we utilize an extended dataset of the grid loss prediction dataset\,\cite{noauthor_grid_nodate}, which covers hourly power consumption and grid loss measurements from May 2017 to August 2023.

While technical grid losses from physical effects like ohmic and corona losses are understood\,\cite{grotmol_robust_2023}, their practical prediction is often inaccurate, and accurate measurements are not available or delayed to be included for a prediction. 
This inaccuracy is not due to a lack of models, but rather due to deficiencies in the data itself. 
Faulty sensors, human error, and other unrecorded factors introduce significant errors, which undermine prediction performance.

\begin{figure}[ht]
    \centering
    \includegraphics[width=1\linewidth]{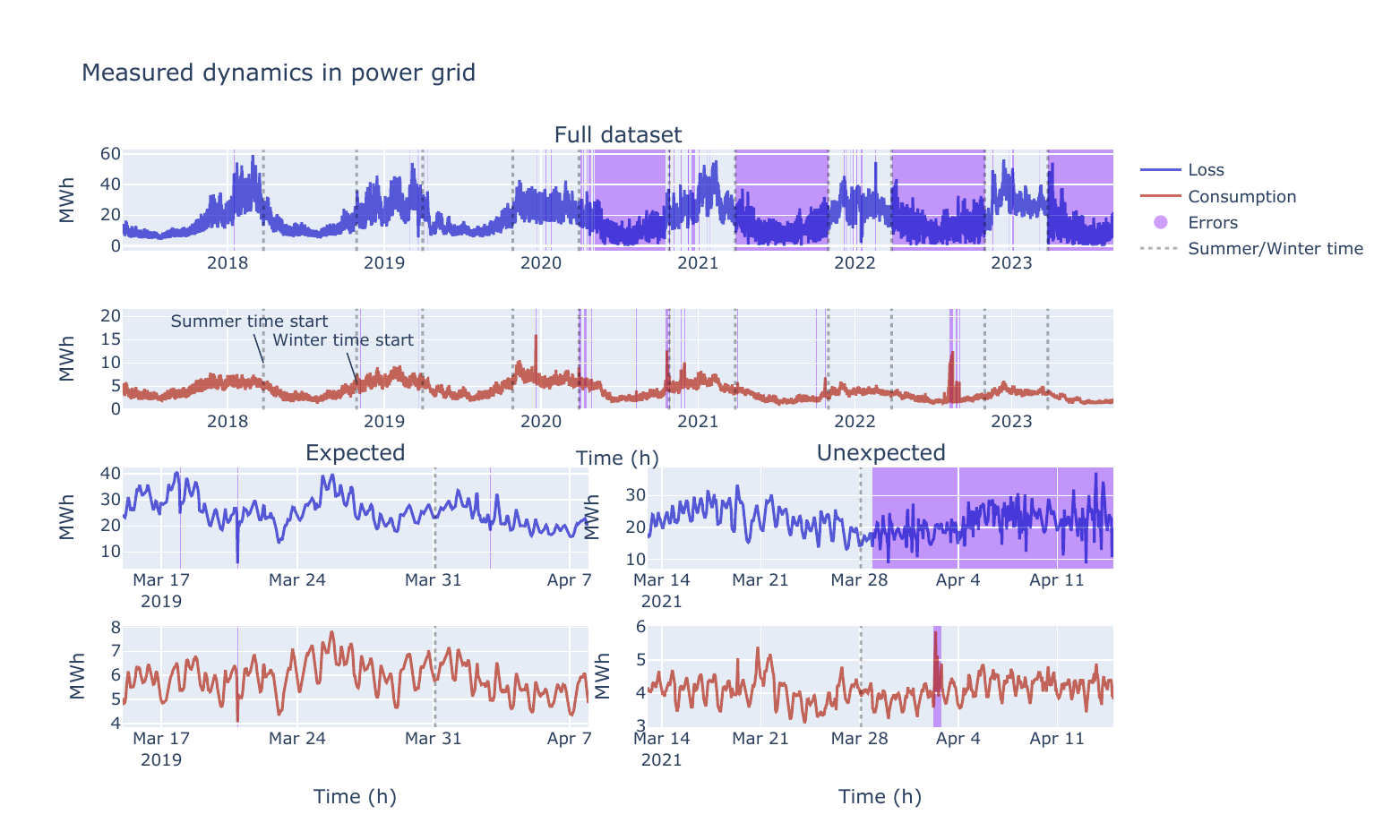}
    \caption{Comparing expected (left) and unexpected (right) behavior in power grid measurements, showing both grid loss and power consumption. The top row displays the full time series, while the bottom row zooms in on specific sections for a closer look. The expected behavior on the left is mostly normal with a few unusual spikes. In contrast, the behavior on the right starts normally but then shifts to consistently unusual readings, a change that coincides with the start of daylight saving time.}
    \label{fig:aneo-dataset-overview}
\end{figure}

The severity of this data quality issue is evident in the grid loss measurements, as shown in \cref{fig:aneo-dataset-overview} with the marked error sections. 
In contrast to the relatively clean power consumption data, the grid loss signal exhibits increased noise levels, particularly during the summer months following April 2020. 
These extended error periods, manually flagged by inspection, begin and end in alignment with daylight saving time changes, as visible in the two plots on the bottom right of \cref{fig:aneo-dataset-overview}.
This systematic noise in the grid loss data is the central problem we aim to address, as cleaner data is essential for improving the accuracy of day-ahead loss predictions.

We divide the dataset into three main sections to validate our approach. The first section acts as the training set where we apply CINDI to iteratively detect and impute errors as detailed in \cref{sec:CINDI-framework}. The second section functions as an evaluation set for ranking candidate solutions during model selection. After improving the data quality of the first section, we use the third section solely as a test set to measure downstream anomaly detection performance. This process fulfills the primary objective of CINDI by producing the clean data required to train robust downstream models\,\cite{xiao_unsupervised_2024,riaz_robust_2025}.

We create four different sets of training data $\mathcal{D}_1, \ldots, \mathcal{D}_4$ with increasing levels of errors, starting at $0\%$ and ending at $24.19\%$, such that
$\mathcal{D}_1\subseteq \cdots \subseteq \mathcal{D}_4$,
where $\mathcal{D}_4$ is the largest training set and each smaller set is a subset of all subsequent larger sets.
\cref{tab:aneo-dataset-split-by-times} summarizes the date ranges for each partition and shows the different levels of errors present in each set.

\begin{table}[h]
	\centering
	\caption{Overview of the dataset splits, detailing their time ranges and the percentage of data points flagged as errors.}
	\label{tab:aneo-dataset-split-by-times}
	\begin{tabular}{lllrr}
	Name            & Start Date   & End Date         & Error \%  & Length \\ 
	\midrule
	$\mathcal{D}_1$ Train (0\%)     & \multirow{4}{5em}{2017-05-01}   & 2019-03-17       & 0.00\%    & 16461 \\ 
	$\mathcal{D}_2$ Train (1.04\%)  &    & 2020-05-03       & 1.04\%    & 26370 \\
	$\mathcal{D}_3$ Train (13.69\%) &    & 2020-10-15       & 13.69\%   & 30330 \\
	$\mathcal{D}_4$ Train (24.19\%) &    & 2021-10-21       & 24.19\%   & 39233 \\
	\midrule
	Evaluation      & 2022-01-06   & 2022-07-01       & 56.72\%   & 4224 \\
	Test            & 2022-11-01   & 2023-08-25       & 52.25\%   & 7149 \\ 
	\end{tabular}
\end{table}

\subsection{Synthetic Evaluation Data}
\label{subsec:fsb-evaluation-data}

For a controlled validation of our approach, we use the Fully Synthetic Benchmark suite (FSB) from the mTADS repository\,\cite{baumgartner_mtads_2023}.
The FSB provides a testbed of 70 synthetic sequences with a range of base signals and different types of anomalies.
This setup allows us to evaluate the model's performance on problems where the data and anomalies are fully defined and controlled.

FSB contains per synthetic sequence a training sequence with and without anomalies, plus a sequence for testing with anomalies.
Essentially, we are using the training sequences with anomalies for training and the sequence without anomalies as the evaluation sequence. 
This means we have no anomaly labels for the evaluation sequence and can not utilize the main objective function from \cref{eq:aneo-cma-es-objective}.
Instead, we use the objective in \cref{eq:fsb-cma-es-objective} for model selection.

\section{Experiments}
\label{sec:results}

In this section, we present experiments to evaluate the performance of the proposed CINDI framework. 
The code and experiment results are available online
\footnote{See \url{https://github.com/2er0/CINDI}}.
All experiments were conducted on a computing cluster equipped with various GPUs, ranging from P100 to H100.

CINDI, described in \cref{sec:CINDI-framework}, can be applied to improve a dataset.
The enhanced dataset is subsequently employed in a specific downstream task, utilizing either CINDI's conditional normalizing flow or an alternative method.
Here, we focus on anomaly detection with CINDI, aiming to identify marked errors in the test set after enhancing the training dataset. 
We adopt the definition for anomaly detection from \citet{chandola_anomaly_2009}, referring to it as \textit{``the problem of finding patterns in data that do not conform to expected or normal behavior''}. 
This definition emphasizes the importance of understanding expected behavior to distinguish patterns that do not align with these expectations and is also relevant for error imputations.
We report the results using F1, VUS, and AUC metrics on the test set.
Both VUS and AUC are non-parametric, and the F1 threshold is set based on the intersection of the AUC-ROC curve with the diagonal.

\subsection{Setup}
\label{subsec:experiment-setup}

The experiment setup for the results is as follows, and we test the imputation fulfillment indirectly via the anomaly detection downstream task. 
We normalize each set of data (train, evaluation, test) based on the training data from which we derive the normalization factors. 
The training data is divided into training and validation sets in a $80:20$ split, where five random sections are selected and expanded to make up 20\% for validation.
CINDI is tested against several baseline methods for imputing the errors, which are treated as missing values in this case. 
We use the following methods\,\footnote{See \url{https://pandas.pydata.org/docs/reference/api/pandas.DataFrame.interpolate.html}}: 'cubic', 'cubicspline', 'linear', 'nearest', 'quadratic', and 'slinear'.
We further test with and without errors in the training data, referred to as 'raw' and 'skip', respectively. 
In addition, we test with the following model-based imputation methods: 'dynamix'\,\cite{hemmer_true_2025} and 'knowimp'\,\cite{chen_rethinking_2024}. 
With model-based methods, we impute one error section at a time, providing each method with a temporal context of 150 days to predict an imputation of the required length.

We use the conditional normalizing flow in CINDI for the anomaly detection task independently of the imputation method; therefore, we test the benefit of improving the training data indirectly.
Using CINDI's strategy for model selection in every scenario ensures fair anomaly detection capabilities, even with different versions of the imputed dataset.
For the final downstream task, we utilize the model selection function described in \cref{subsec:cindi-model-selection}, omitting the reconstruction term to focus solely on anomaly detection.

For the conditional normalizing flow, we test with three different encoder types to utilize the temporal context: 'base', 'MLP', and 'CNN'.
The base version does no encoding and passes the temporal context information on without processing.
The MLP and CNN versions utilize a single model to encode the temporal context before passing it on to all transformation layers, employing an amortized approach.

Results in Tab.\,\ref{tab:aneo-results-overview} are based on the model selection process. 
\cref{fig:aneo-results_1.04} further contains in the box plots the range of performances discovered during this search.
The reported max performance is derived from all steps on the training dataset, including imputation and detection.

\subsection{Results: Grid Loss Data}
\label{subsec:experiment-results-aneo}

The analysis of the results highlighted that CINDI can improve a dataset up to a certain percentage of errors, in our case, up to $13.69\%$, as shown in \cref{tab:aneo-results-overview}, which summarizes all downstream results.
For CINDI, we show both final and maximum performance across iterations. 
This demonstrates that any model evaluated during the iterative process can be tested for its downstream performance, sometimes performing well even if not explicitly targeted for that task.
For all other methods, the results reflect the best model identified through the model selection process.

We do not report results on the training set $\mathcal{D}_1$ with $0\%$ errors in \cref{tab:aneo-results-overview}, as no imputation is needed in this case and only the anomaly detection task is relevant. 
For these cases, we provide the average and standard deviation across all experiments. 
The average results are: F1 score of $89\% \pm 4\%$, VUS of $93\% \pm 5\%$, and AUC of $93\% \pm 5\%$.

\begin{table}[ht]
    \centering
    \caption{Overview of test results on the grid loss data with anomaly detection as a downstream task. With the CINDI methods, we present the final performance of the last iteration and, in brackets, the maximum performance reached during all iteration steps. Values in \textbf{bold} highlight the highest performing score on the given dataset with a specific metric. Not completed experiments are shown as '-'. Dynamix as proposed by \cite{hemmer_true_2025} and KnowIpm by \cite{chen_rethinking_2024}.}
    \label{tab:aneo-results-overview}
    \vspace{0.3em}
    \resizebox{0.8\textwidth}{!}{
        \scriptsize
        \begin{tabular}{ll|lll|lll|lll}
			& & \multicolumn{3}{|l}{$\mathcal{D}_2$ Train 1.04\,\%} & \multicolumn{3}{|l}{$\mathcal{D}_3$ Train 13.69\,\%} & \multicolumn{3}{|l}{$\mathcal{D}_4$ Train 24.19\,\%} \\
			& & F1 & VUS & AUC & F1 & VUS & AUC & F1 & VUS & AUC \\
			Method & Encoder & final & final & final & final & final & final & final & final & final \\
			& & (max) & (max) & (max) & (max) & (max) & (max) & (max) & (max) & (max) \\
			\midrule
			CINDI & Base & \textbf{0.93} & 0.97 & 0.97 & 0.92 & \textbf{0.97} & 0.96 & 0.81 & 0.82 & 0.81 \\
			& & \textbf{(0.94)} & (0.97) & (\textbf{0.98}) & (0.92) & \textbf{(0.97)} & (0.96) & (0.86) & (0.89) & (0.88) \\
			CINDI & CNN & 0.83 & 0.83 & 0.82 & 0.82 & 0.85 & 0.85 & 0.65 & 0.76 & 0.74 \\
			& & (0.91) & (0.96) & (0.95) & (0.84) & (0.88) & (0.87) & (0.81) & (0.86) & (0.85) \\
			CINDI & MLP & 0.90 & 0.94 & 0.96 & 0.75 & 0.83 & 0.82 & 0.80 & 0.74 & 0.72 \\
			& & (0.91) & (0.96) & (0.96) & (0.86) & (0.91) & (0.91) & (0.85) & (0.88) & (0.88) \\
			\midrule
			\midrule
			 & & F1 & VUS & AUC & F1 & VUS & AUC & F1 & VUS & AUC \\
			\midrule
			Nearest & Base & 0.74 & 0.79 & 0.77 & \textbf{0.94} & 0.96 & \textbf{0.97} & 0.87 & 0.94 & 0.93 \\
			Nearest & CNN & 0.83 & 0.87 & 0.87 & 0.88 & 0.90 & 0.91 & \textbf{0.94} & 0.96 & \textbf{0.97} \\
			Nearest & MLP & 0.83 & 0.77 & 0.76 & 0.78 & 0.81 & 0.80 & 0.92 & 0.96 & \textbf{0.97} \\
			Linear & Base & 0.74 & 0.78 & 0.77 & 0.80 & 0.88 & 0.87 & 0.88 & 0.94 & 0.94 \\
			Linear & CNN & 0.78 & 0.84 & 0.83 & 0.91 & 0.94 & 0.96 & 0.89 & 0.94 & 0.94 \\
			Linear & MLP & 0.87 & 0.91 & 0.90 & 0.88 & 0.89 & 0.89 & 0.90 & 0.96 & 0.96 \\
			Slinear & Base & 0.89 & 0.86 & 0.90 & 0.87 & 0.93 & 0.92 & 0.81 & 0.88 & 0.88 \\
			Slinear & CNN & 0.92 & 0.91 & 0.95 & 0.92 & 0.88 & 0.96 & 0.92 & 0.96 & 0.96 \\
			Slinear & MLP & 0.85 & 0.89 & 0.88 & 0.87 & 0.86 & 0.86 & 0.90 & 0.96 & 0.96 \\
			Quadratic & Base & 0.89 & 0.93 & 0.94 & 0.80 & 0.85 & 0.84 & 0.87 & 0.93 & 0.93 \\
			Quadratic & CNN & 0.83 & 0.84 & 0.84 & 0.82 & 0.88 & 0.88 & 0.90 & 0.96 & \textbf{0.97} \\
			Quadratic & MLP & 0.76 & 0.84 & 0.83 & 0.89 & 0.85 & 0.84 & 0.88 & 0.94 & 0.94 \\
			Cubic & Base & 0.74 & 0.81 & 0.80 & 0.83 & 0.88 & 0.86 & 0.90 & 0.95 & 0.95 \\
			Cubic & CNN & 0.75 & 0.77 & 0.76 & 0.86 & 0.92 & 0.92 & 0.89 & 0.95 & 0.95 \\
			Cubic & MLP & 0.82 & 0.90 & 0.89 & 0.82 & 0.80 & 0.78 & 0.88 & 0.94 & 0.94 \\
			Cubicspline & Base & 0.78 & 0.73 & 0.71 & 0.00 & 0.51 & 0.50 & 0.00 & 0.51 & 0.50 \\
			Cubicspline & CNN & 0.80 & 0.77 & 0.76 & 0.00 & 0.49 & 0.46 & 0.00 & 0.52 & 0.54 \\
			Cubicspline & MLP & 0.85 & 0.91 & 0.91 & 0.00 & 0.51 & 0.50 & 0.00 & 0.52 & 0.54 \\
			\midrule
			Skip & Base & \textbf{0.93} & 0.97 & 0.97 & 0.91 & 0.96 & 0.95 & 0.86 & 0.94 & 0.93 \\
			Skip & CNN & 0.87 & 0.92 & 0.91 & 0.93 & 0.96 & 0.96 & 0.93 & \textbf{0.97} & \textbf{0.97} \\
			Skip & MLP & 0.84 & 0.91 & 0.90 & 0.83 & 0.90 & 0.89 & 0.90 & 0.96 & 0.96 \\
			Raw & Base & \textbf{0.93} & \textbf{0.98} & \textbf{0.98} & 0.88 & 0.90 & 0.90 & 0.84 & 0.83 & 0.82 \\
			Raw & CNN & 0.75 & 0.77 & 0.76 & 0.86 & 0.91 & 0.92 & 0.91 & 0.95 & 0.95 \\
			Raw & MLP & 0.88 & 0.90 & 0.90 & 0.81 & 0.82 & 0.80 & 0.78 & 0.68 & 0.66 \\
			\midrule
			Dynamix & Base & 0.77 & 0.74 & 0.72 & 0.87 & 0.93 & 0.93 & 0.91 & 0.92 & 0.96 \\
			Dynamix & CNN & 0.73 & 0.73 & 0.74 & 0.92 & 0.88 & 0.96 & 0.91 & 0.91 & 0.96 \\
			Dynamix & MLP & 0.91 & 0.96 & 0.95 & 0.90 & 0.94 & 0.94 & 0.91 & 0.91 & 0.96 \\
			KnowImp & Base & 0.92 & 0.87 & 0.96 & 0.88 & 0.92 & 0.92 & - & - & - \\
			KnowImp & CNN & 0.85 & 0.85 & 0.85 & 0.86 & 0.86 & 0.85 & - & - & - \\
			KnowImp & MLP & 0.83 & 0.68 & 0.66 & 0.80 & 0.81 & 0.80 & - & - & - \\
			\end{tabular}
			
    }
\end{table}

\begin{figure}[ht]
    \centering
    \includegraphics[width=1\linewidth]{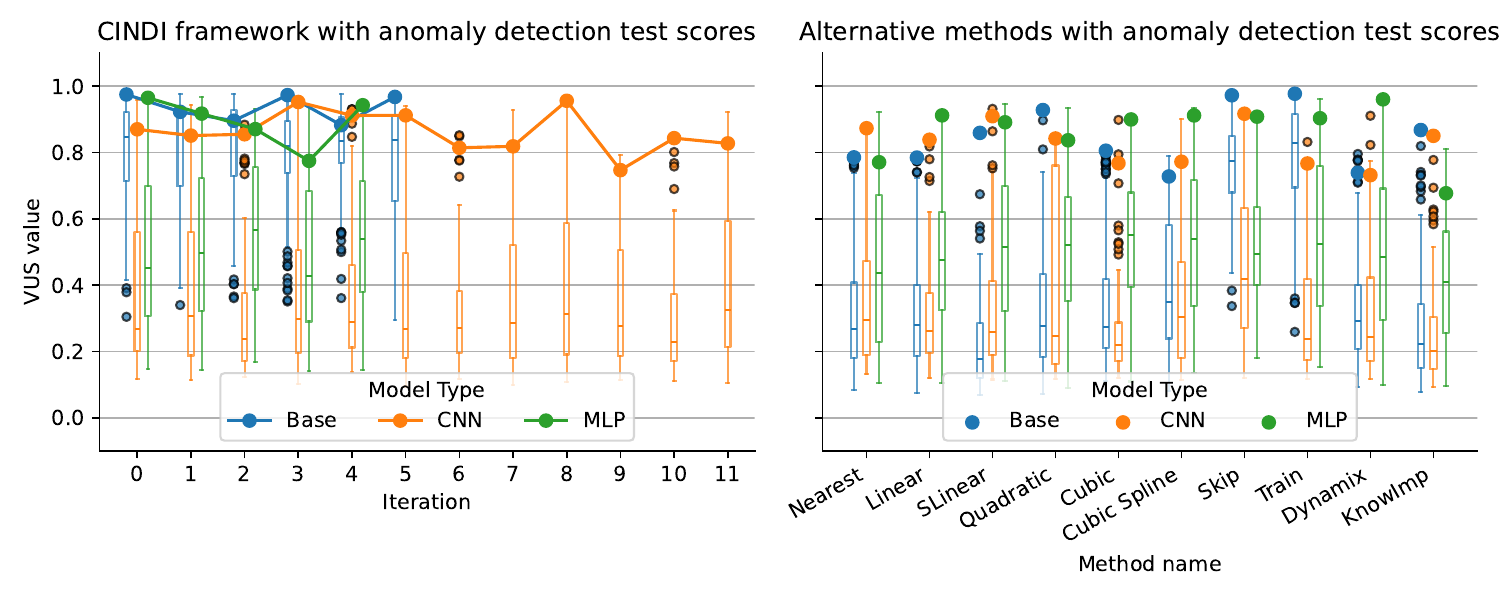}
    \caption{VUS performance results for CINDI and baselines with $1.04\%$ errors in the training data. Points are the final model performance after model selection, and box plots show all the tested candidate solutions.}
    \label{fig:aneo-results_1.04}
\end{figure}

\cref{fig:aneo-results_1.04} shows the results at an error level of $1.04\%$.
Further figures for higher error levels ($13.69\%$ and $24.19\%$) and F1 scores are provided in Appendix\,\ref{apx:aneo-result-plots-remaining} and in our repository.
In each plot, the solid lines (left) and scatter points (right) represent the anomaly detection performance using the selected model. 
In contrast, the box plots show the performance ranges of all evaluated candidates.
This demonstrates that the model selection process explores many options and that the chosen model achieves good performance on the evaluation set and the test set.

\begin{figure}[ht]
\centering
\begin{tabular}{cc}
 \includegraphics[width=0.5\linewidth]{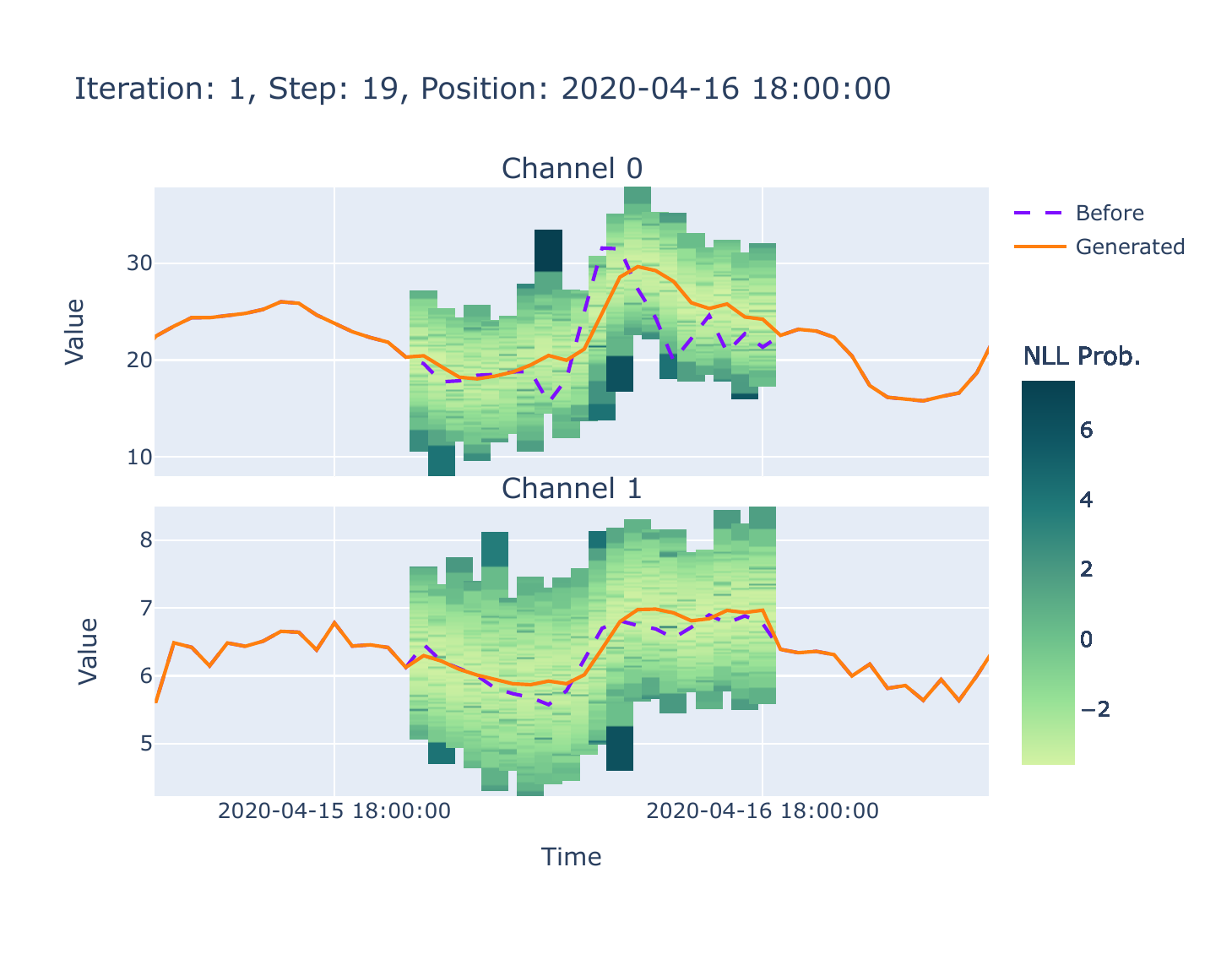} &  \includegraphics[width=0.5\linewidth]{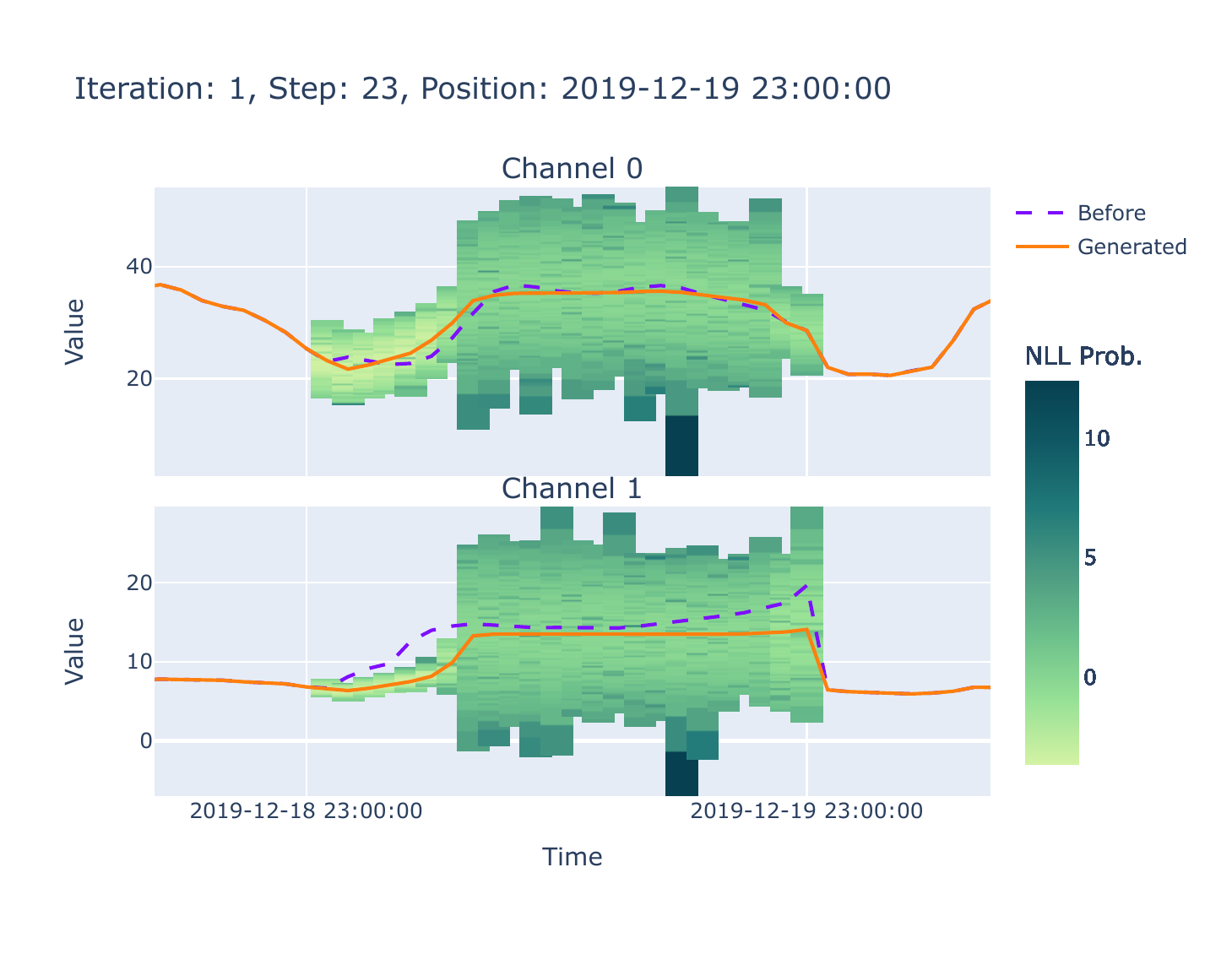} \\
{\scriptsize (a) Imputation example 1} & {\scriptsize (b) Imputation example 2} \\
\end{tabular}
\caption{Results from the second iteration on the dataset with $1.04\%$ noise. Fig.\,(a), (b) show self-regressing imputation of two flagged sections, where the heatmap indicates the negative log-likelihood of possible other samples.}
\label{fig:base-1-percent-imputation-overview}
\end{figure}

\begin{figure}[ht]
\centering
\begin{tabular}{c}
 \includegraphics[width=0.8\linewidth]{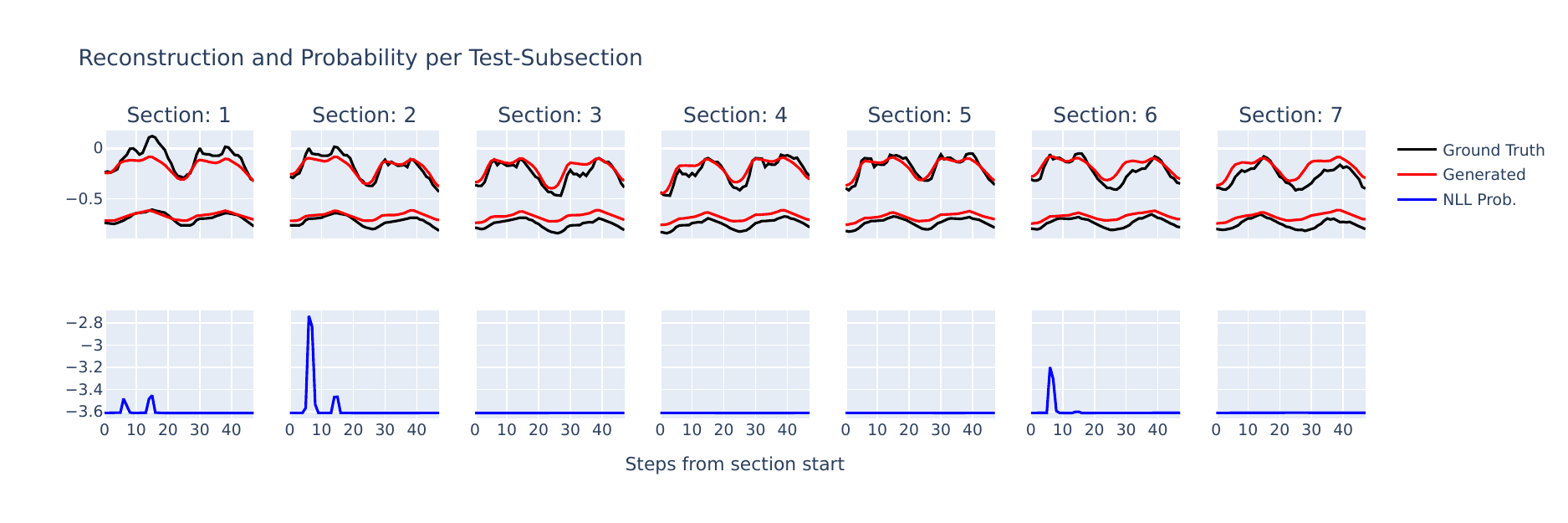} \\
 {\scriptsize (a) Reconstruction performance} \\
 \includegraphics[width=0.8\linewidth]{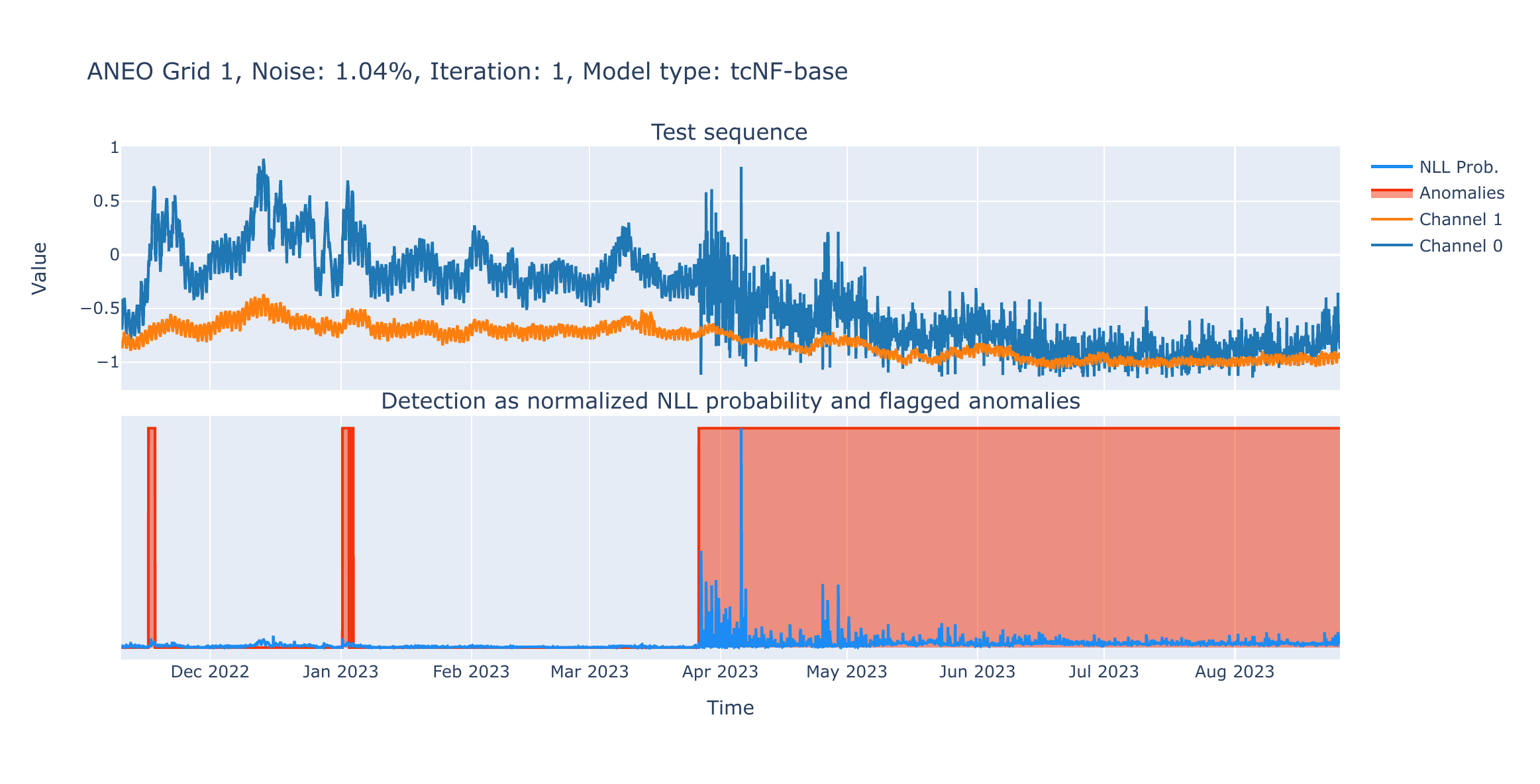} \\
 {\scriptsize (b) Test data and negative log-likelihood} \\
 \includegraphics[width=0.7\linewidth]{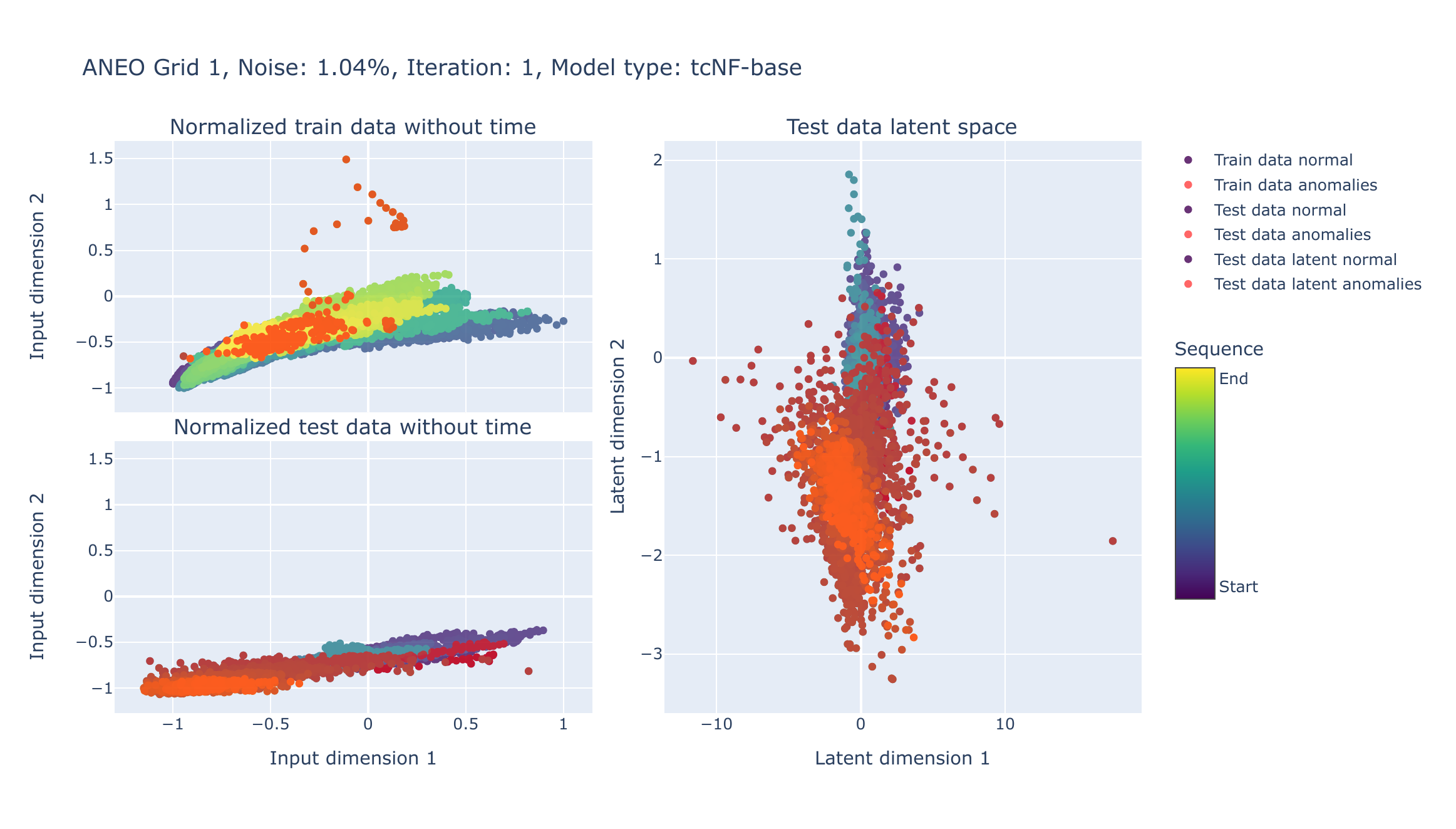} \\
 {\scriptsize (c) Input and latent space} \\
\end{tabular}
\caption{Results from the second iteration on the dataset with $1.04\%$ noise. Fig.\,(a), (b) show self-regressing imputation of two flagged sections, with a heatmap indicating the negative log-likelihood of possible options. Fig.\,(c) shows the reconstruction of expected data with its negative log-likelihood. Fig.\,(d) and (e) display test data, detected anomalies, and latent space.}
\label{fig:base-1-percent-detection-overview}
\end{figure}

\cref{fig:base-1-percent-imputation-overview} and \cref{fig:base-1-percent-detection-overview} show CINDI's imputation in the second iteration on the training data.
\cref{fig:base-1-percent-imputation-overview}(a) and (b) show two flagged sections being imputed. 
Where \cref{fig:base-1-percent-imputation-overview}(a) results in natural-looking sequences, while plot (b) reveals some uncertainty, especially after the first few steps.
In \cref{fig:base-1-percent-imputation-overview}(a), the heatmap displays the most likely imputation path, which is less clear in \cref{fig:base-1-percent-imputation-overview}(b) with the increased uncertainty.

\cref{fig:base-1-percent-detection-overview}(a) shows the reconstruction of seven two-day long sections.
Here, the model generally captures expected behavior, with only minor deviations, such as the omission of slight variations in complex regions.
Overall, the reconstruction closely matches the data, therefore performing well in self-regression forecasting.

\cref{fig:base-1-percent-detection-overview}(b) and (c) show the anomaly detection downstream task if this specific model had been chosen as the final one.
While it performs well (F1 score $0.87$, VUS score $0.92$), it does not outperform the best model found for this task (F1 score $0.93$, VUS score $0.97$) in this end-to-end execution of CINDI.

\begin{figure}
\begin{tabular}{cc}
  \includegraphics[width=0.5\linewidth]{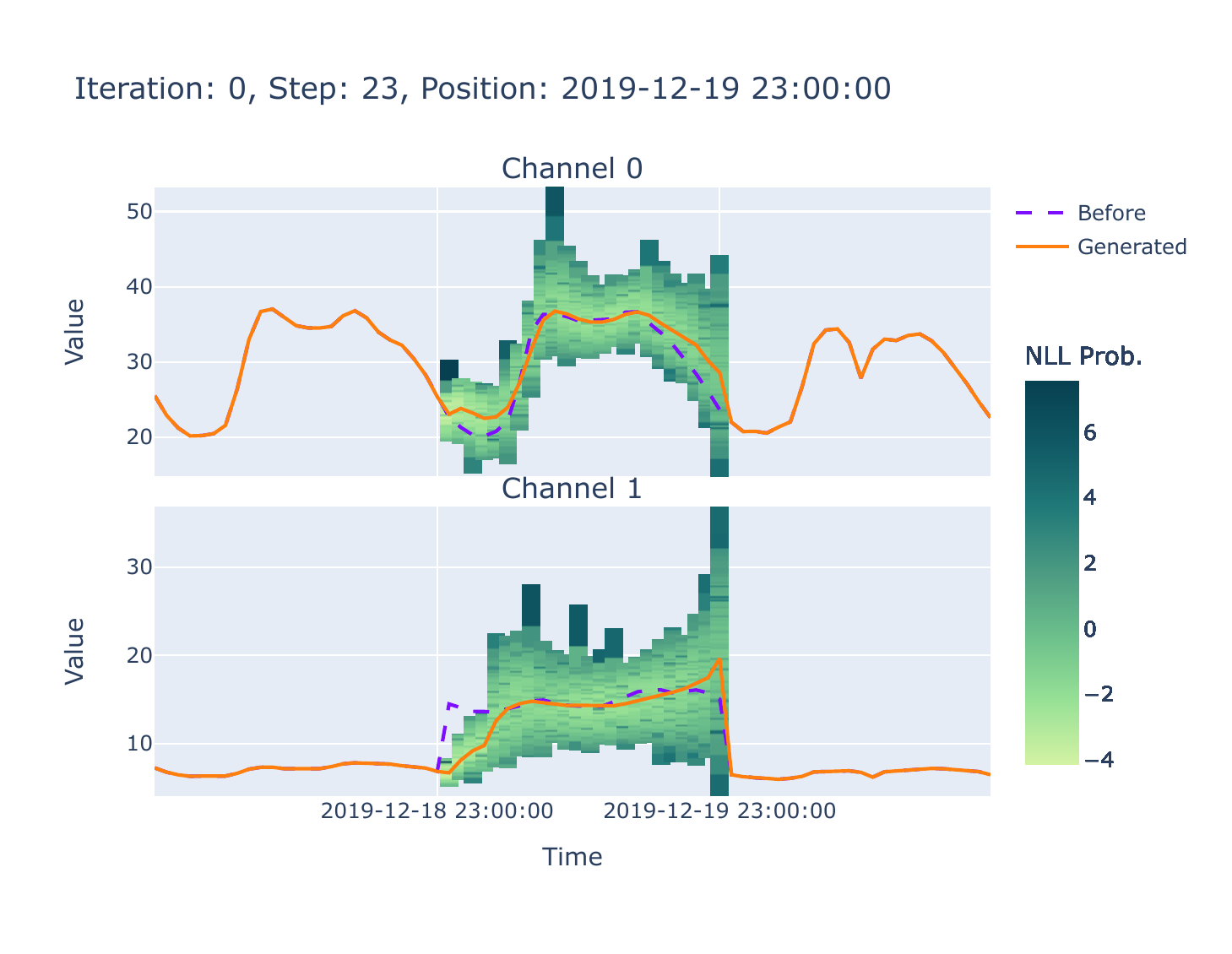} &   \includegraphics[width=0.5\linewidth]{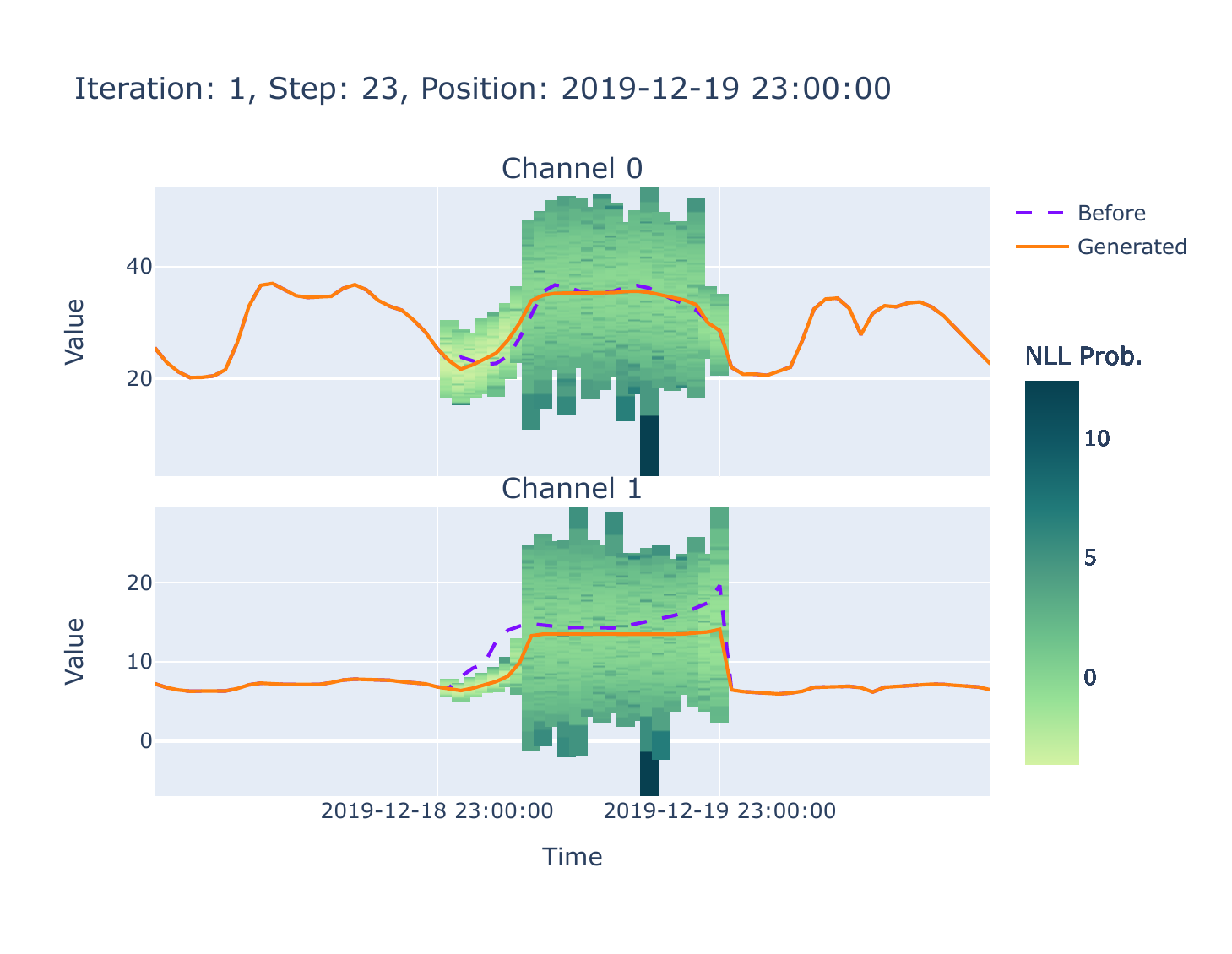} \\
{\scriptsize (a) 1st iteration} & {\scriptsize (b) 2nd iteration} \\
 \includegraphics[width=0.5\linewidth]{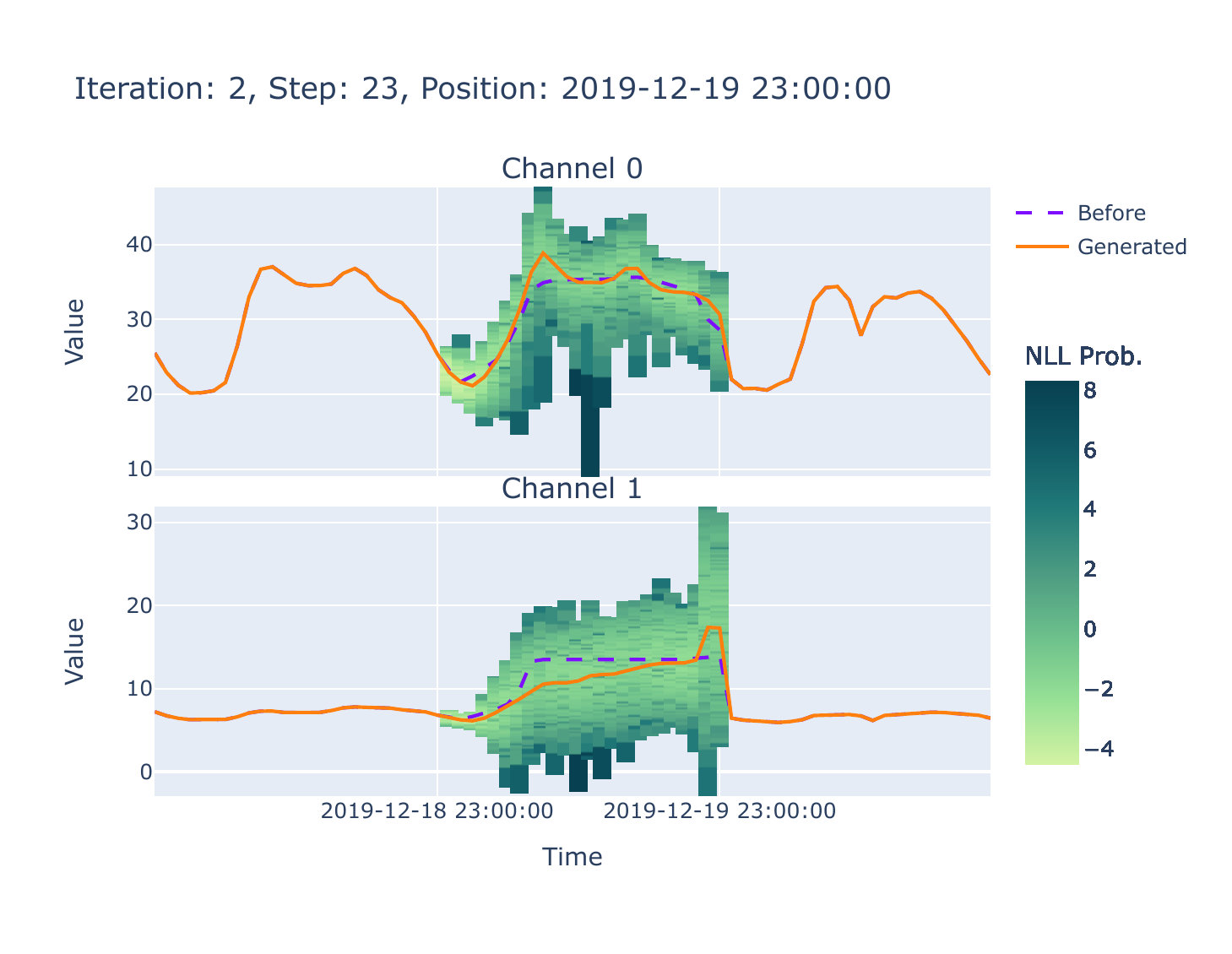} &   \includegraphics[width=0.5\linewidth]{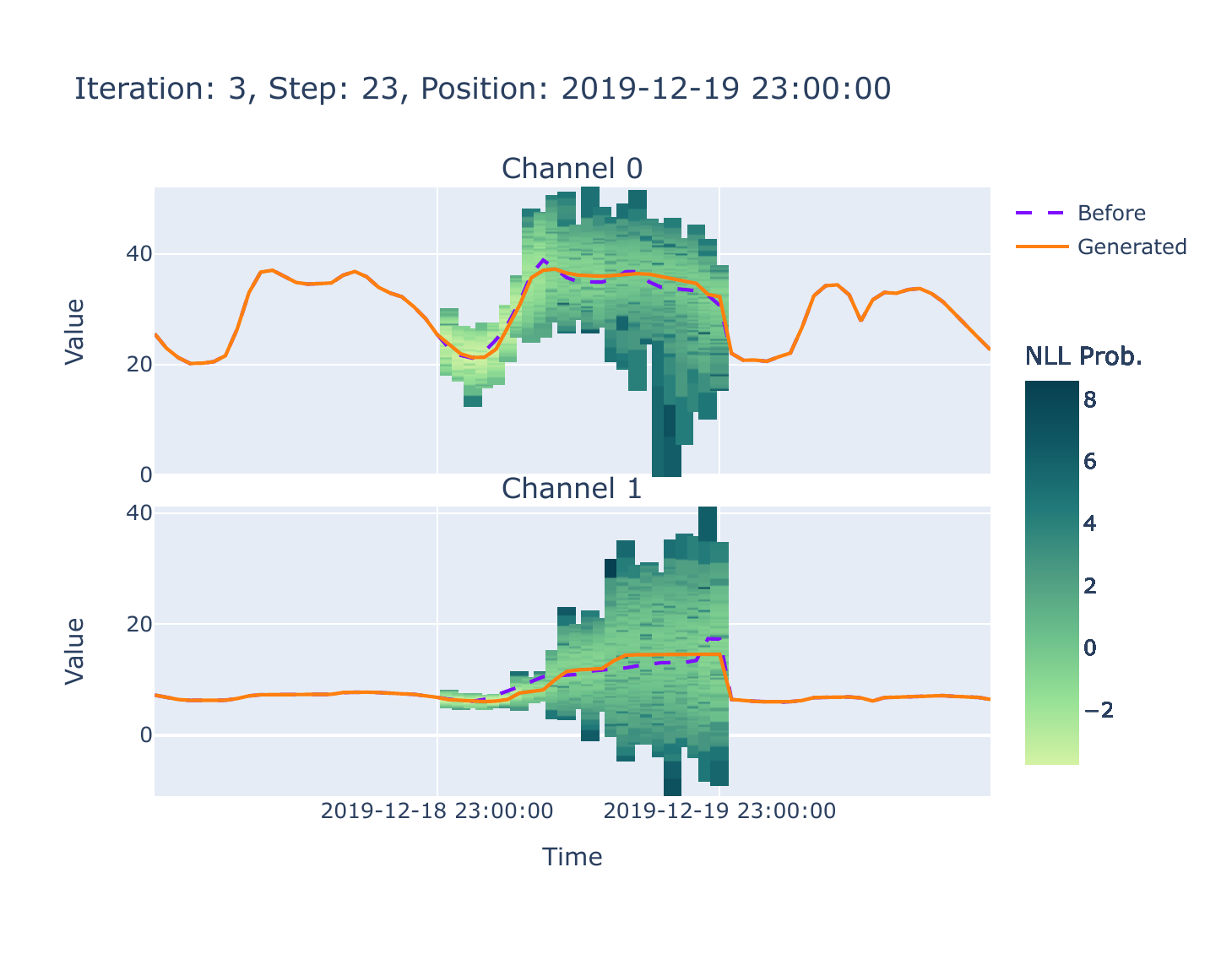} \\
{\scriptsize (c) 3rd iteration} & {\scriptsize (d) 4th iteration} \\
\end{tabular}
\caption{Imputation of one section across the first four iterations with CINDI on the dataset with $1.04\%$ noise.}
\label{fig:base-1-percent-one-section-four-steps}
\end{figure}

\begin{figure}
    \centering
    \includegraphics[width=1\linewidth]{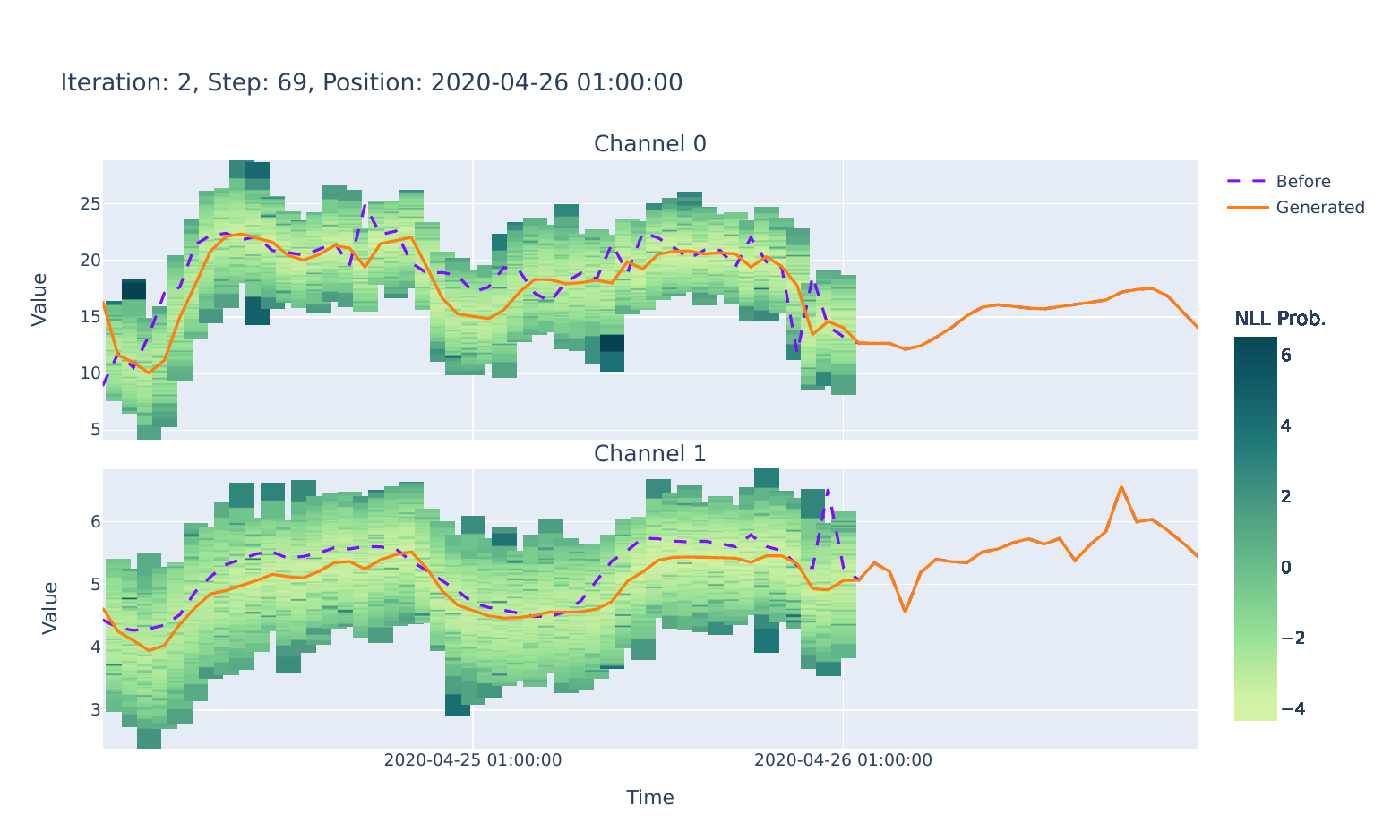}
    \caption{Imputation of an extended marked section in the third iteration with CINDI.}
    \label{fig:base-1-percent-long-impute}
\end{figure}

The collection in \cref{fig:base-1-percent-one-section-four-steps} tracks the process of one flagged section being imputed over four iterations.
This demonstrates how CINDI refines its imputations until it settles on a final replacement, which is an improvement over the original sequence, though not always natural and intuitively desired by an observer.
This highlights the system's flexibility in continuing to update already imputed sections as needed.
With \cref{fig:base-1-percent-long-impute}, we show that CINDI can also handle longer error sections, imputing sequences beyond two days without introducing unexpected artifacts of this extended period.

\cref{tab:aneo-results-overview} further demonstrates the impact of increasing error levels in the training data. 
More errors reduce the benefits of imputing error sections with CINDI, resulting in decreased performance. 
It also shows that simply skipping error or marked sections generally does not harm the downstream performance and allows for solid performances.
Further, the pretrained model 'dynamix'\,\cite{hemmer_true_2025} leads to solid performance with increasing level of errors in the training data and shows no performance degradation.
This approach utilizes the pre-trained model to impute all error sections, meaning it has been pre-trained and therefore does not need to rely on unclean data to begin with.

\subsection{Results: FSB}
\label{subsec:experiment-results-fsb}

On the FSB datasets, CINDI imputes training data errors under specific conditions: the data must contain noise, such as Gaussian noise, to a small percentage and a structure or pattern. 
In the case where the data do not fulfill these requirements, the conditional normalizing flow will learn a new manifold representation of the data, which will not follow the base distribution.
Examples of imputations on the FSB dataset are provided in Appendix\,\ref{apx:fsb-result-collection}.

\section{Limitations \& Remarks}
\label{sec:limitations}

Our CINDI framework is designed as a single unified system that utilizes a single model per iteration while being capable of handling different tasks. 
This simplicity helps keep it efficient, while also encouraging exploration of the limits of a unified solution and the reuse of learned behavior in practical scenarios.

When comparing CINDI to other methods, it is clear that the baselines are highly competitive and can be difficult to outperform in this context. 
For this reason, we ensured that the model selection capabilities of CINDI were utilized for every option involved in the experiments, thereby giving every option the best possible chance.

One interesting finding is that simply skipping ('skip') the error sections is a strong option. 
This means that avoiding imputation altogether for sections with errors can still lead to solid results, and it should not be ignored as a baseline.
This holds especially with increased levels of errors in the training data, since too many errors can lead to biased models.
Standard imputation methods, such as nearest or linear interpolation, turn out to be reasonable choices, especially as the amount of noise increases. 
In these cases, conditional normalizing flows tend to map these input sections to the same position in the latent space, which limits what can be learned from those data points.

A challenge emerges when there is no noise in the data, as observed with many sequences from the fully synthetic benchmark suite. 
In these cases, training the model results in new manifolds that are not useful for detection or imputation. 
This shows that some level of noise or imperfection is essential for guiding CINDI in practical data improvement and detection applications.

\section{Conclusion}
\label{sec:conclusion}

This work focused on the problem of predicting grid loss from the perspective of anomaly detection, which heavily depends on the quality of available data. 
We introduced CINDI, an unsupervised probabilistic framework designed to enhance data quality by imputing errors in multivariate time series.
By utilizing conditioned normalizing flows, our approach provides flexibility and efficiency, supporting multiple tasks related to data improvement and anomaly detection.

With the end-to-end training, we were able to provide a unified framework mainly for detecting and imputing, but not limited to these tasks.
Our experiments demonstrate that the approach is robust and yields competitive results overall when compared to other methods. 
The framework is effective at denoising existing data, but it cannot reconstruct actual values when the underlying data is missing or severely corrupted.
When the level of errors increases, it becomes difficult for any method to deliver reliable imputations, and as a result, produces less useful replacements, except for pre-trained models.

Overall, the proposed framework represents a practical step toward enhancing data quality for grid loss prediction, thanks to its unified and efficient design that supports multiple use cases.
However, further work is required to overcome current limitations and make the approach even more useful.

Future research directions include:

\begin{itemize}
    \item \textbf{Improving the conditioning mechanism:} Develop more sophisticated strategies for conditioning, such as learning a conditional distribution that better reflects the underlying signal on the distribution side.
    \item \textbf{Selective imputation:} Find ways to identify which components or channels need imputation, such that only affected areas are modified while preserving valid data.
    \item \textbf{Adaptive imputation behavior:} Investigate iterative and adaptive techniques that allow gradual improvements instead of full replacement, which should lead to more stable and reliable results over time.
    \item \textbf{Exploring time embeddings:} Examine time embedding methods to capture temporal patterns more effectively in real-world data, possibly using continuous or learned time features rather than separate static channels.
\end{itemize}

\section*{Acknowledgements}

This work has been carried out at the Centre for Research-based Innovation, SFI NorwAI, funded by the Research Council of Norway under grant no. 309834. 
The authors thank Are Løkken Ottesen and Nisha Dalal (Aneo, \url{www.aneo.com}) for their constructive feedback and support. We are especially grateful for providing an extended version of the grid loss prediction dataset.

\section*{Declaration of generative AI and AI-assisted technologies in the manuscript preparation process.}

During the preparation of the manuscript, the authors utilized Google Gemini to enhance the wording and writing. After using this tool/service, the authors reviewed and edited the content as needed and take full responsibility for the content of the published article. 

\FloatBarrier

\bibliographystyle{elsarticle-num-names}
\bibliography{mybibinline}

\FloatBarrier

\appendix
\section{Additional Grid Loss Dataset Results}
\label{apx:aneo-result-plots-remaining}

\begin{figure}[ht]
    \centering
    \includegraphics[width=1\linewidth]{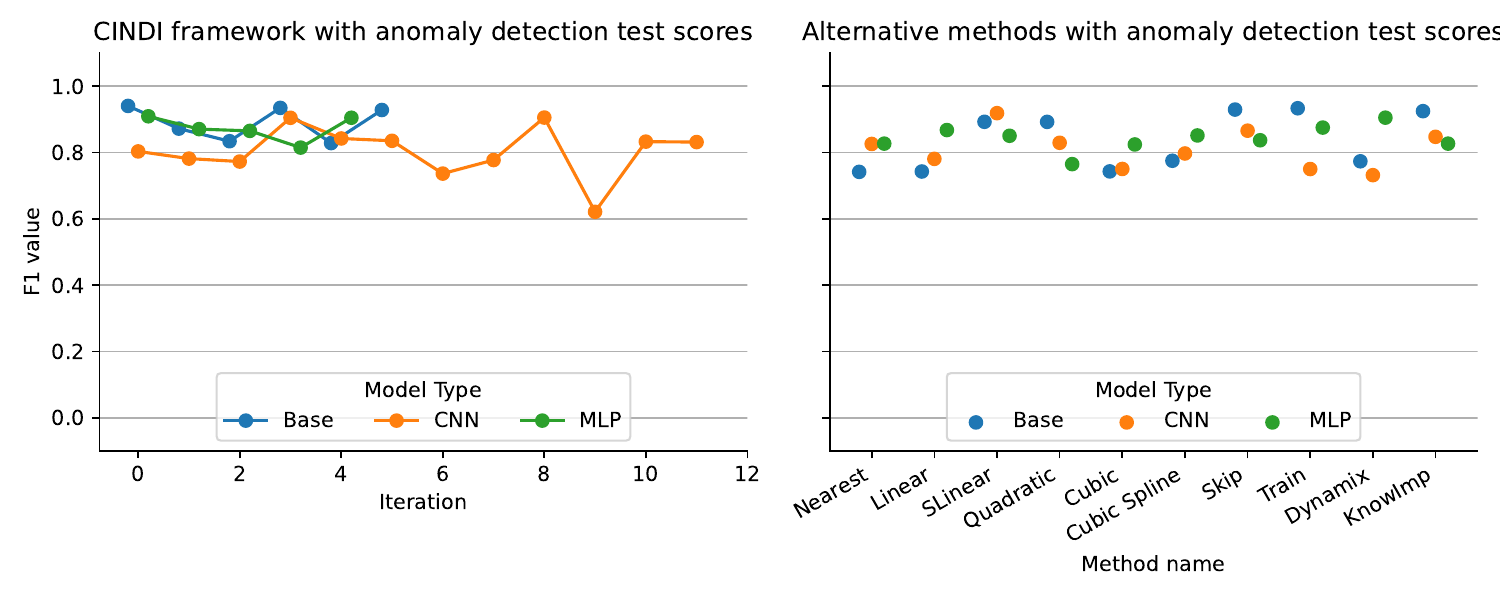}
    \caption{F1 performance results on training dataset with $1.04\%$ noise.}
\end{figure}

\begin{figure}[ht]
    \centering
    \includegraphics[width=1\linewidth]{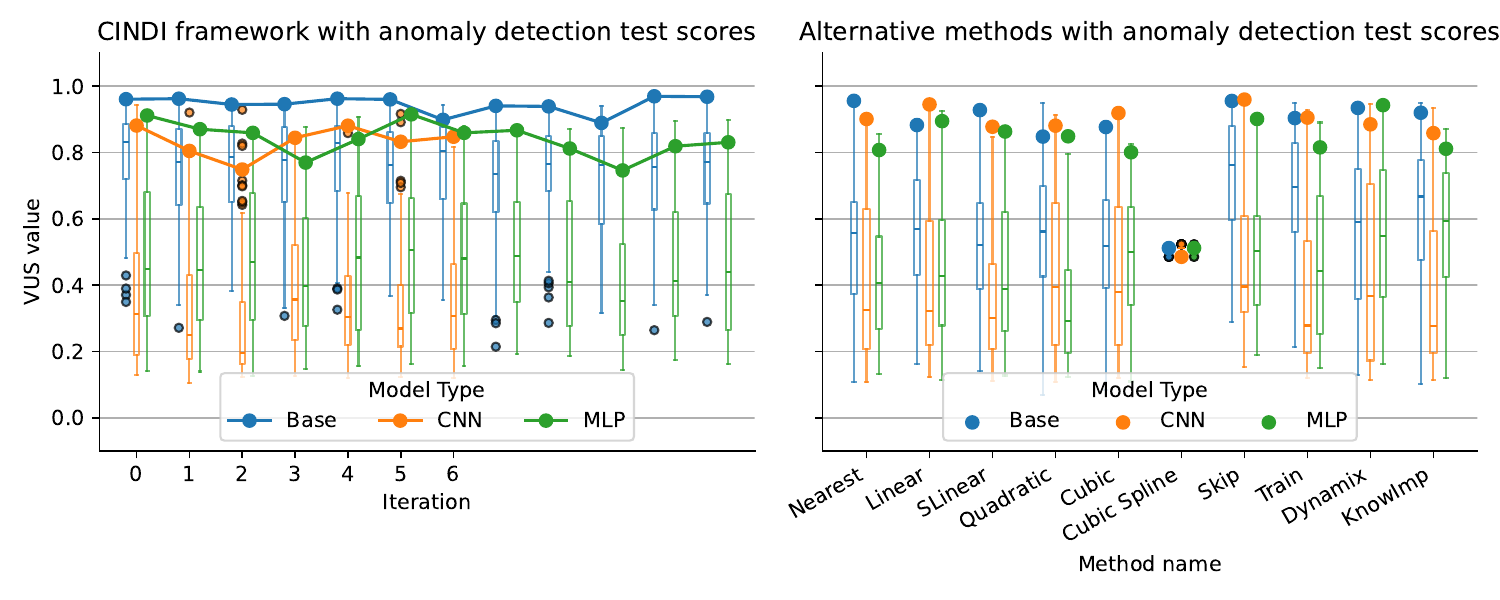}
    \caption{VUS performance results on training dataset with $13.69\%$ noise.}
\end{figure}

\begin{figure}[ht]
    \centering
    \includegraphics[width=1\linewidth]{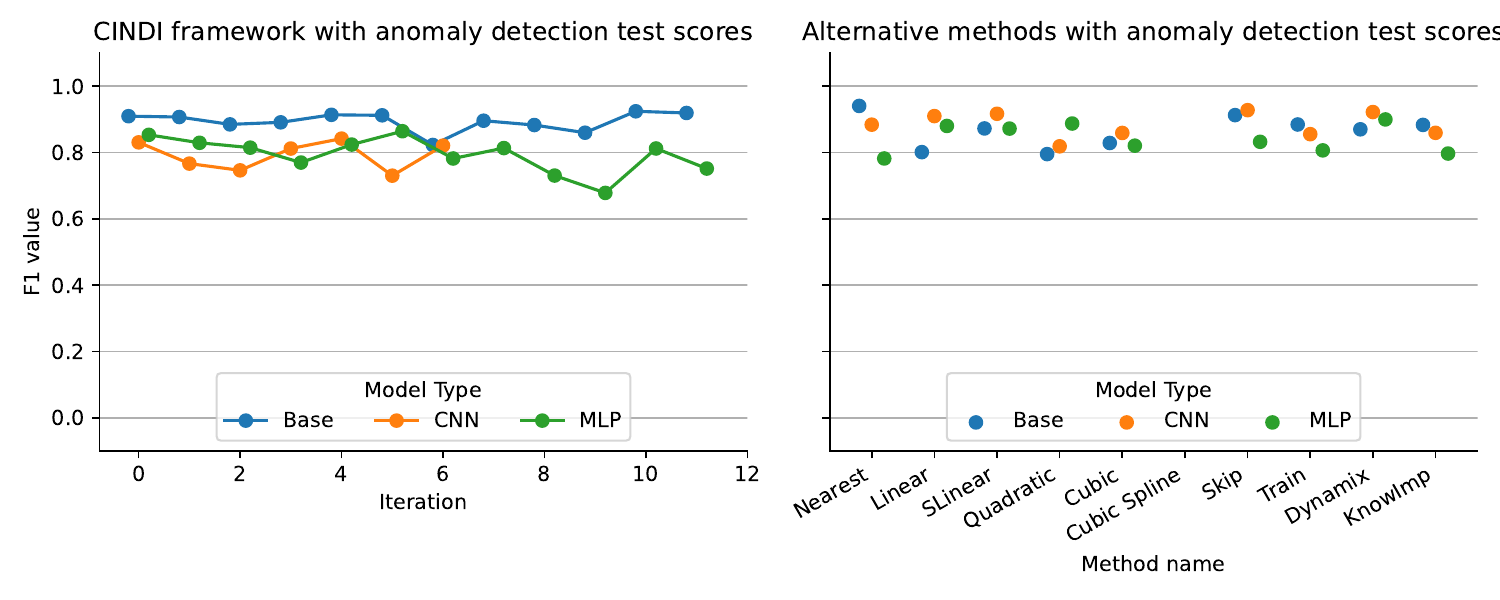}
    \caption{F1 performance results on training dataset with $13.69\%$ noise.}
\end{figure}

\begin{figure}[ht]
    \centering
    \includegraphics[width=1\linewidth]{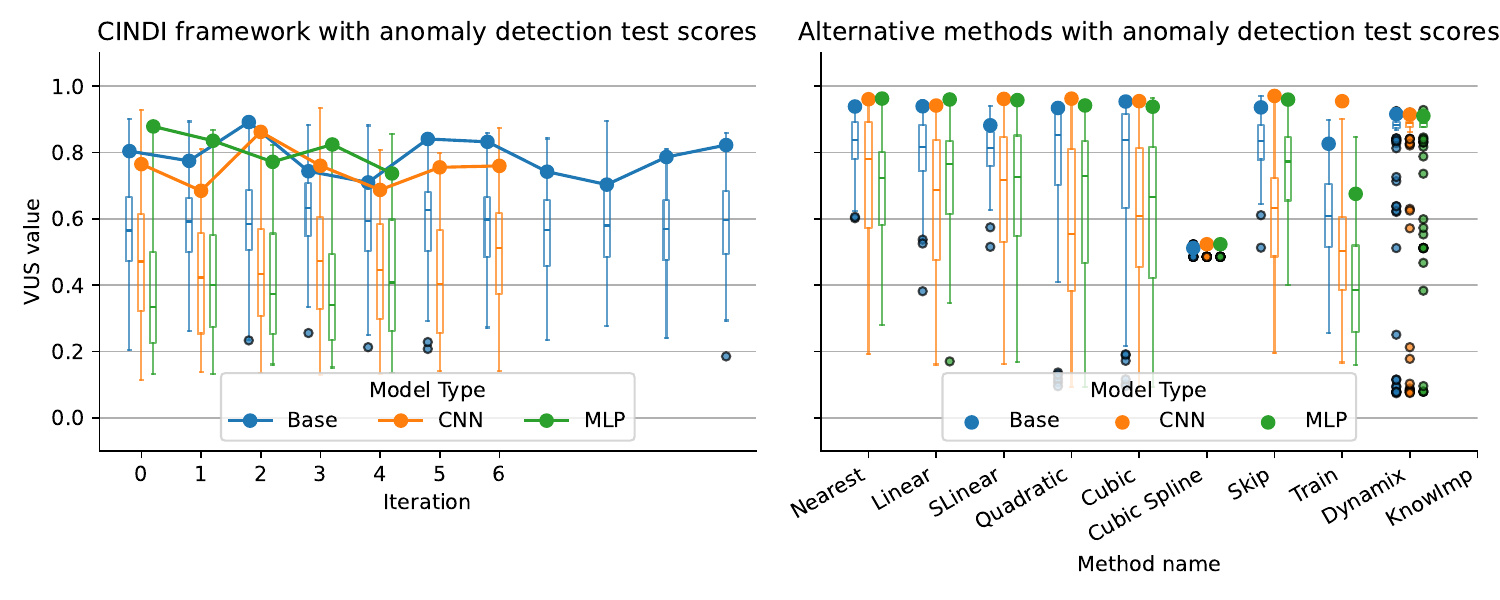}
    \caption{VUS performance results on training dataset with $24.19\%$ noise.}
\end{figure}

\begin{figure}[ht]
    \centering
    \includegraphics[width=1\linewidth]{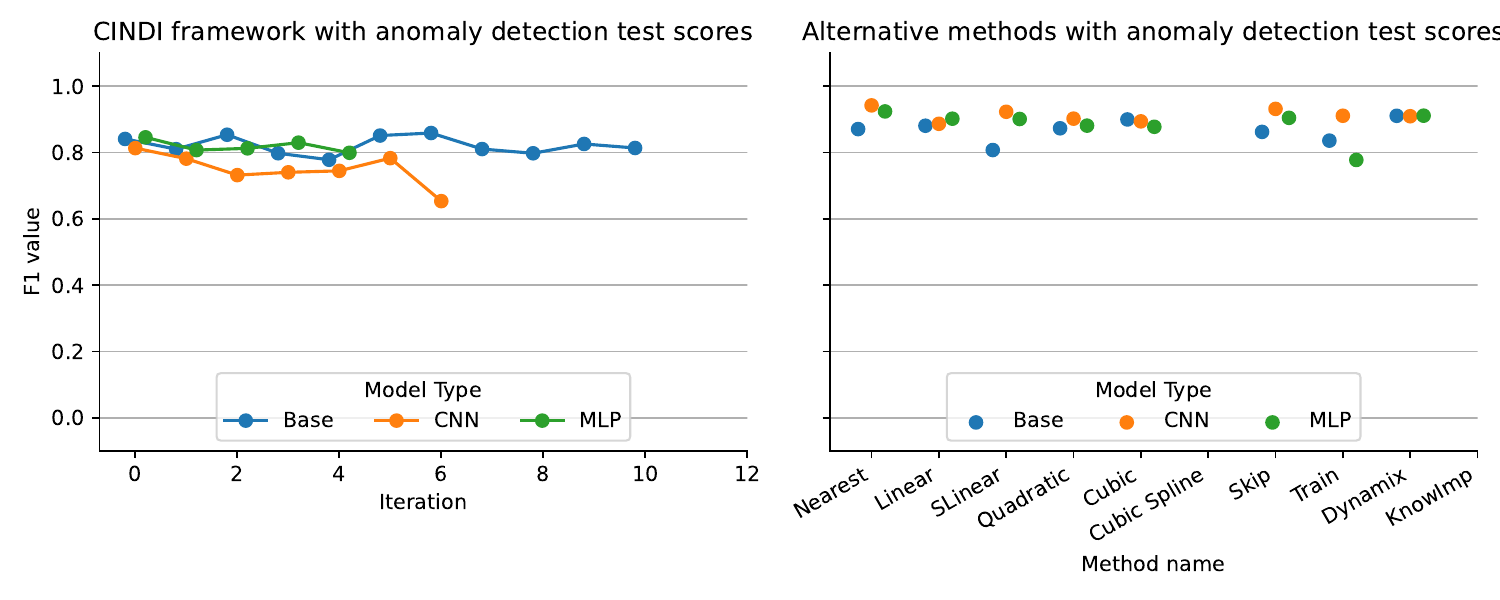}
    \caption{F1 performance results on training dataset with $24.19\%$ noise.}
\end{figure}

\FloatBarrier

\section{FSB Results}
\label{apx:fsb-result-collection}

\subsection{CINDI CNN on 2-sine-all-channel-anomaly Sequence}

\begin{figure}[ht]
\begin{tabular}{cc}
 \includegraphics[width=0.5\linewidth]{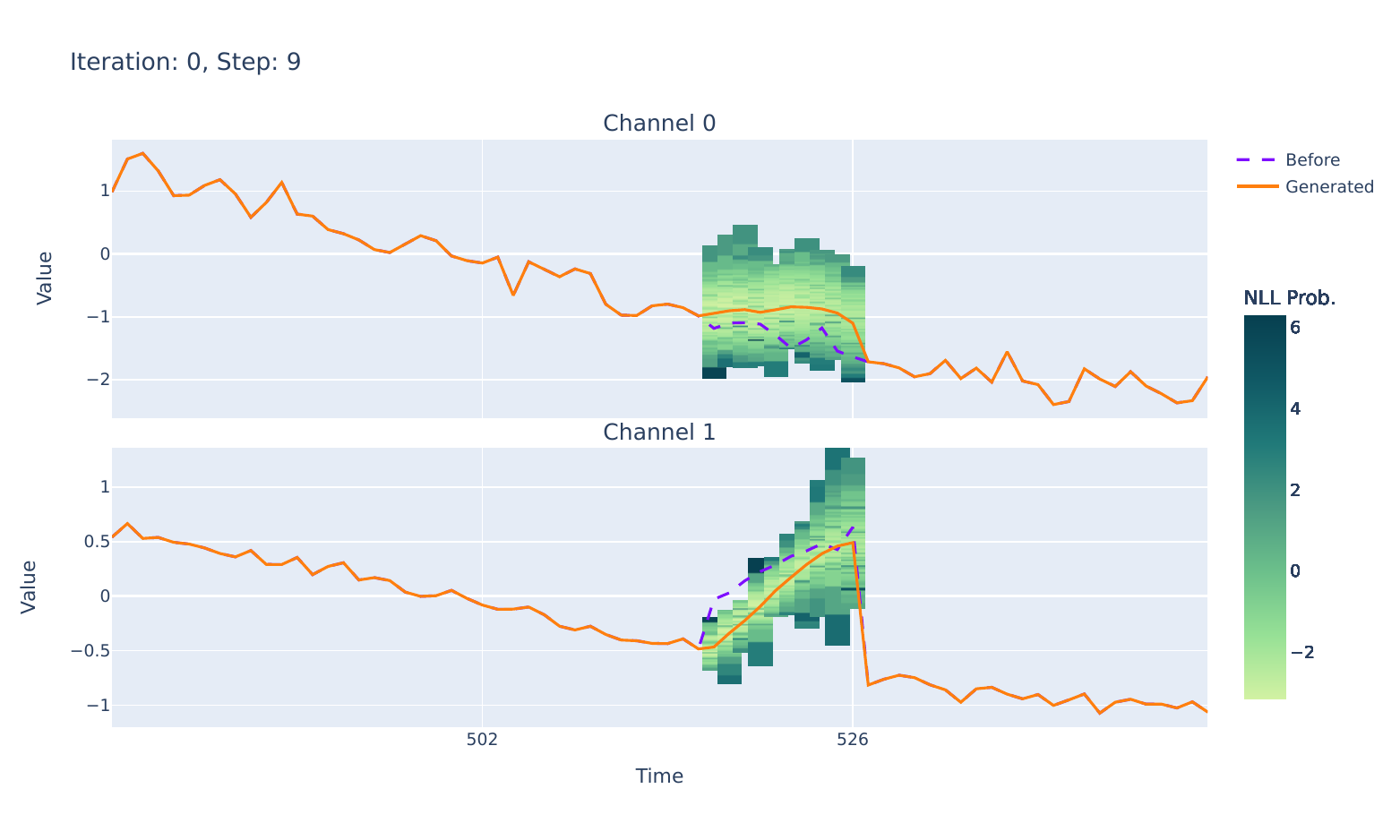} &   \includegraphics[width=0.5\linewidth]{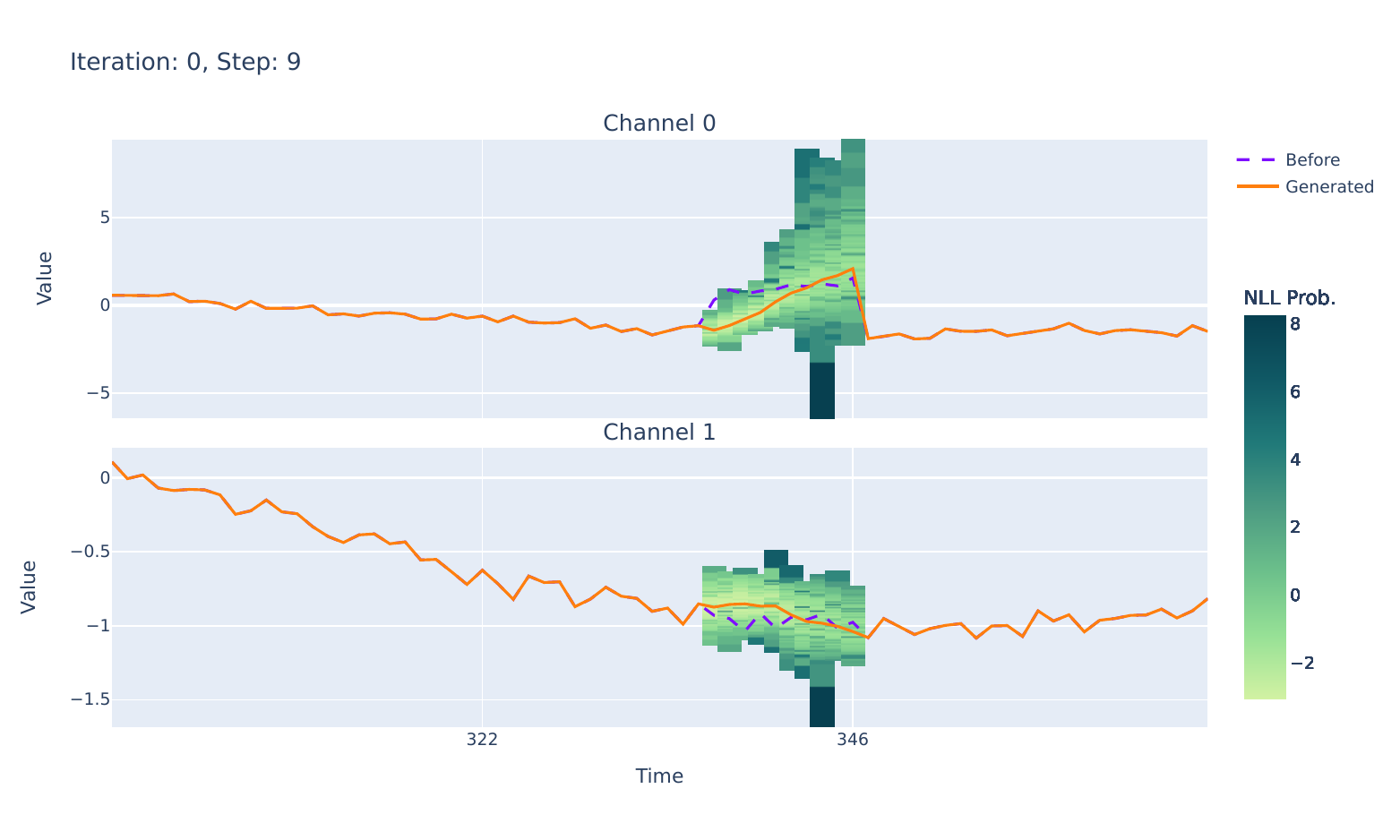} \\
{\scriptsize (a) 1st iteration section 302 to 311} & {\scriptsize (b) 1st iteration section 122 to 131} \\

\includegraphics[width=0.5\linewidth]{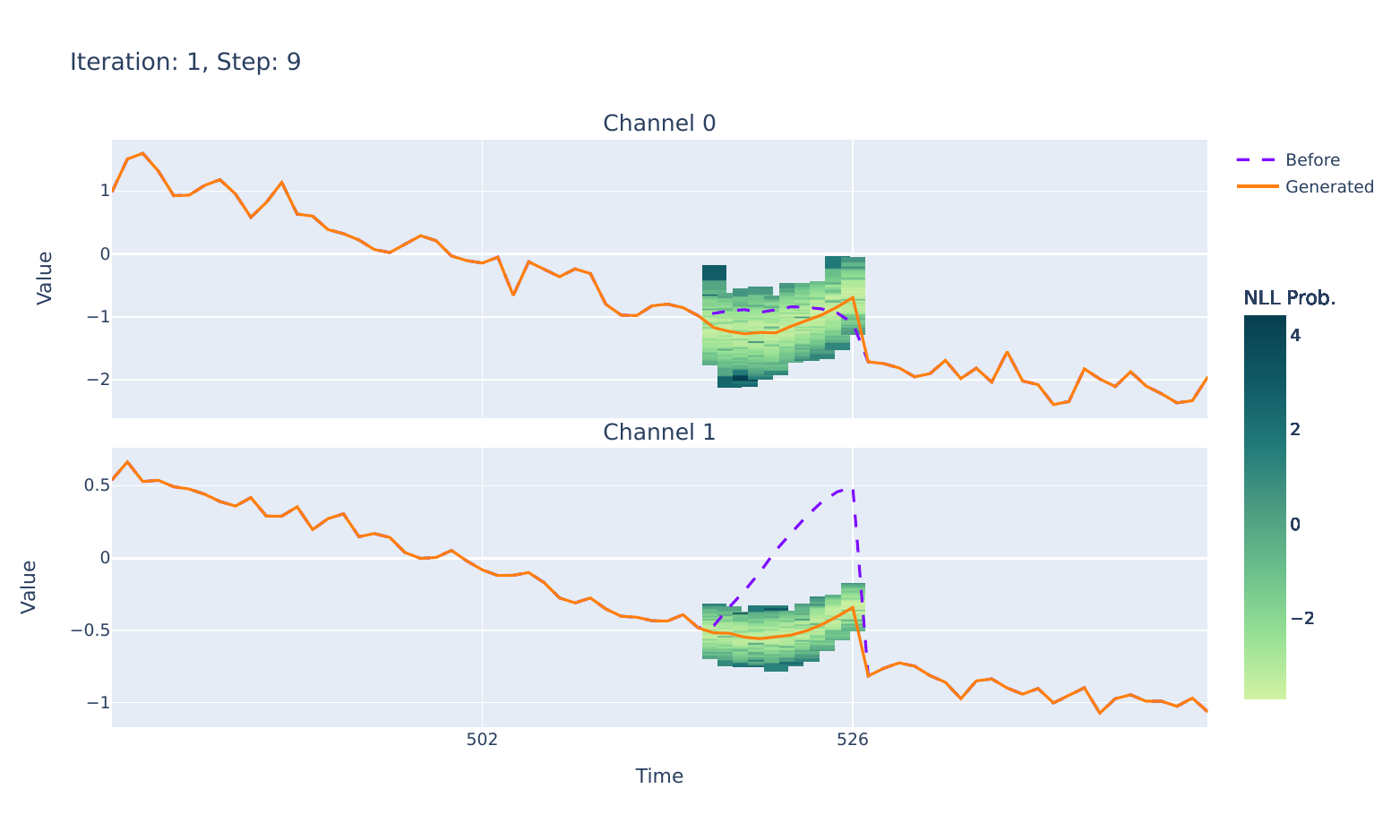} &   \includegraphics[width=0.5\linewidth]{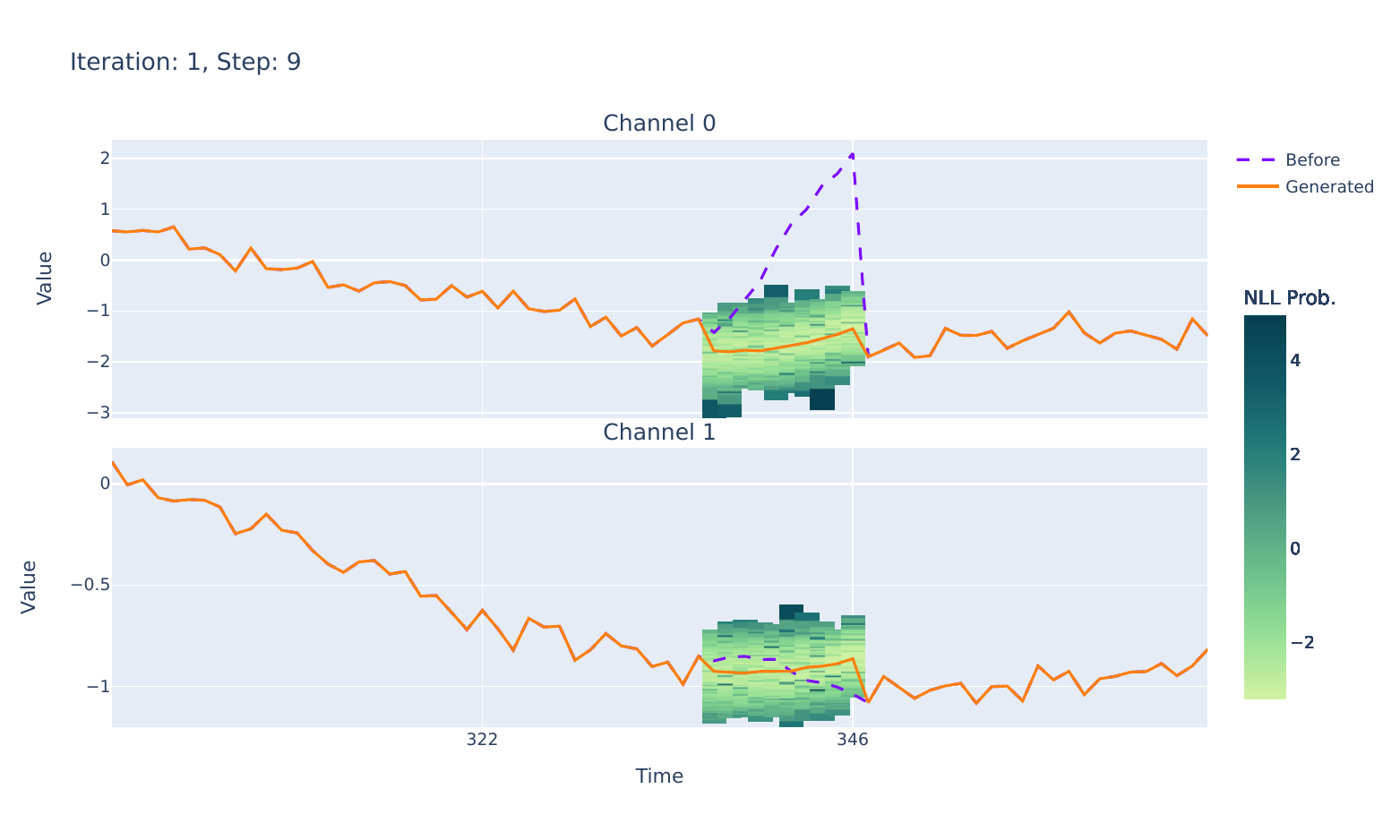} \\
{\scriptsize (c) 2nd iteration section 302 to 311} & {\scriptsize  (d) 2nd iteration section 122 to 131}\\

\includegraphics[width=0.5\linewidth]{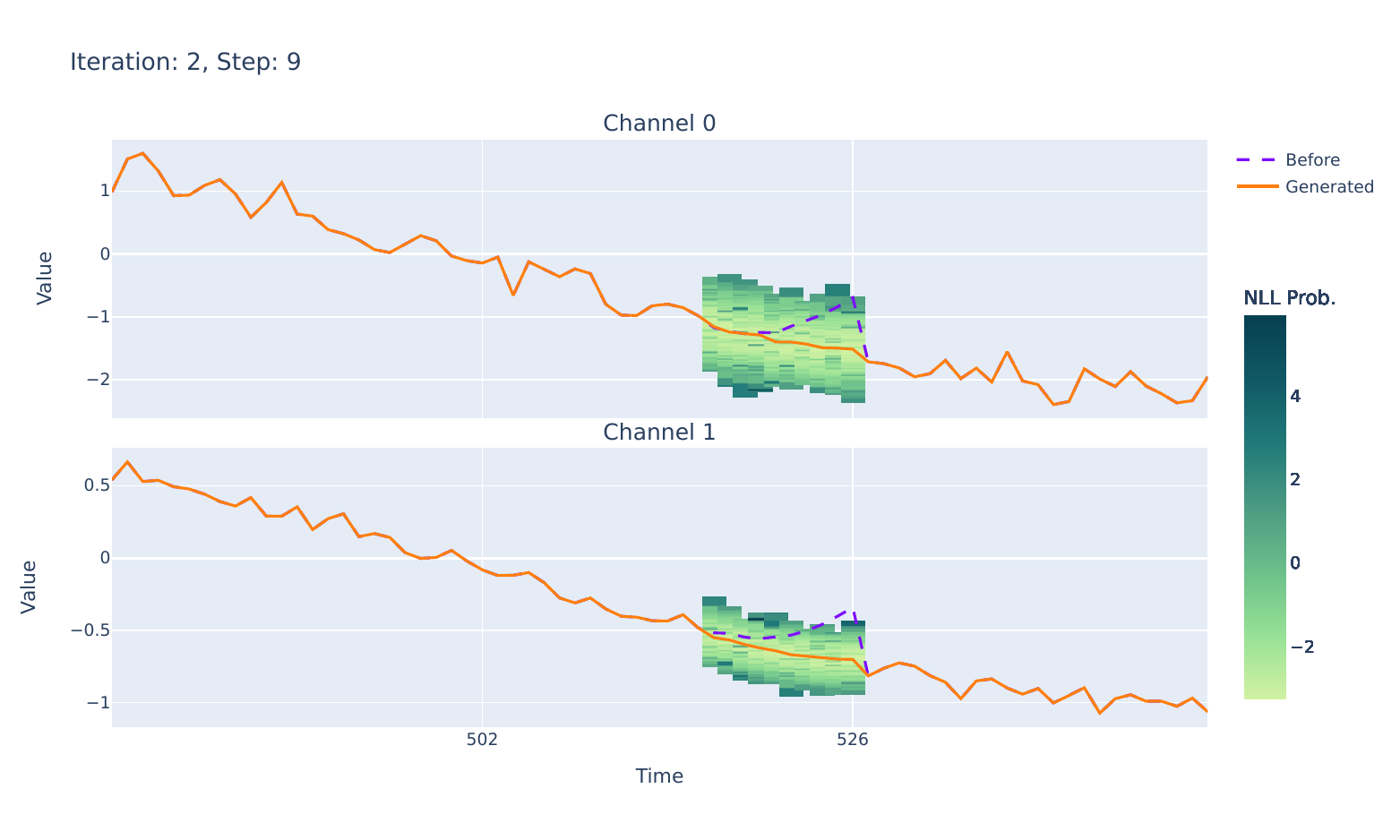} & \includegraphics[width=0.5\linewidth]{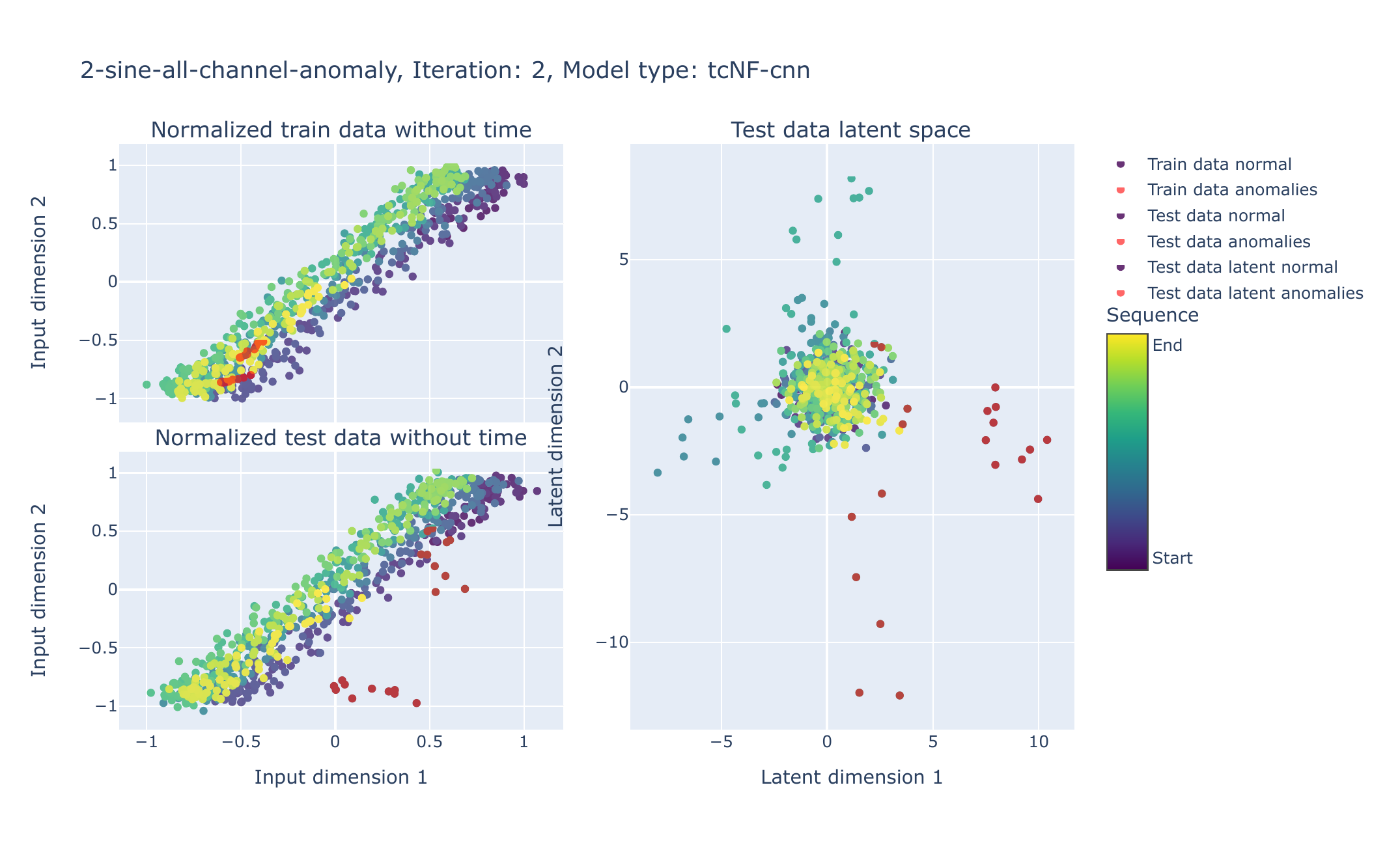}  \\
{\scriptsize (e) 3rd iteration section 302 to 311} & {\scriptsize (f) 3rd iteration input and latent space} \\
 
\end{tabular}
\caption{Imputation of two errors in the training data using CINDI with model type CNN over three iterations, shown in Fig.\,(a)–(e). The anomaly detection performance on the test set across iterations yields F1 scores of $0.33, 0.69, 0.33, 0.44$, and VUS scores of $0.90, 0.96, 0.97, 0.98$. Fig.\,(f) shows the input and latent spaces where the error regions have moved into the main data range. The latent space (right) shows that the error sections in the test set are shifted into lower density areas.}
\end{figure}

\FloatBarrier

\subsection{CINDI CNN on 2-sine-one-channel-anomaly-noise-10\% Sequence}

\begin{figure}[ht]
    \centering
    \includegraphics[width=1\linewidth]{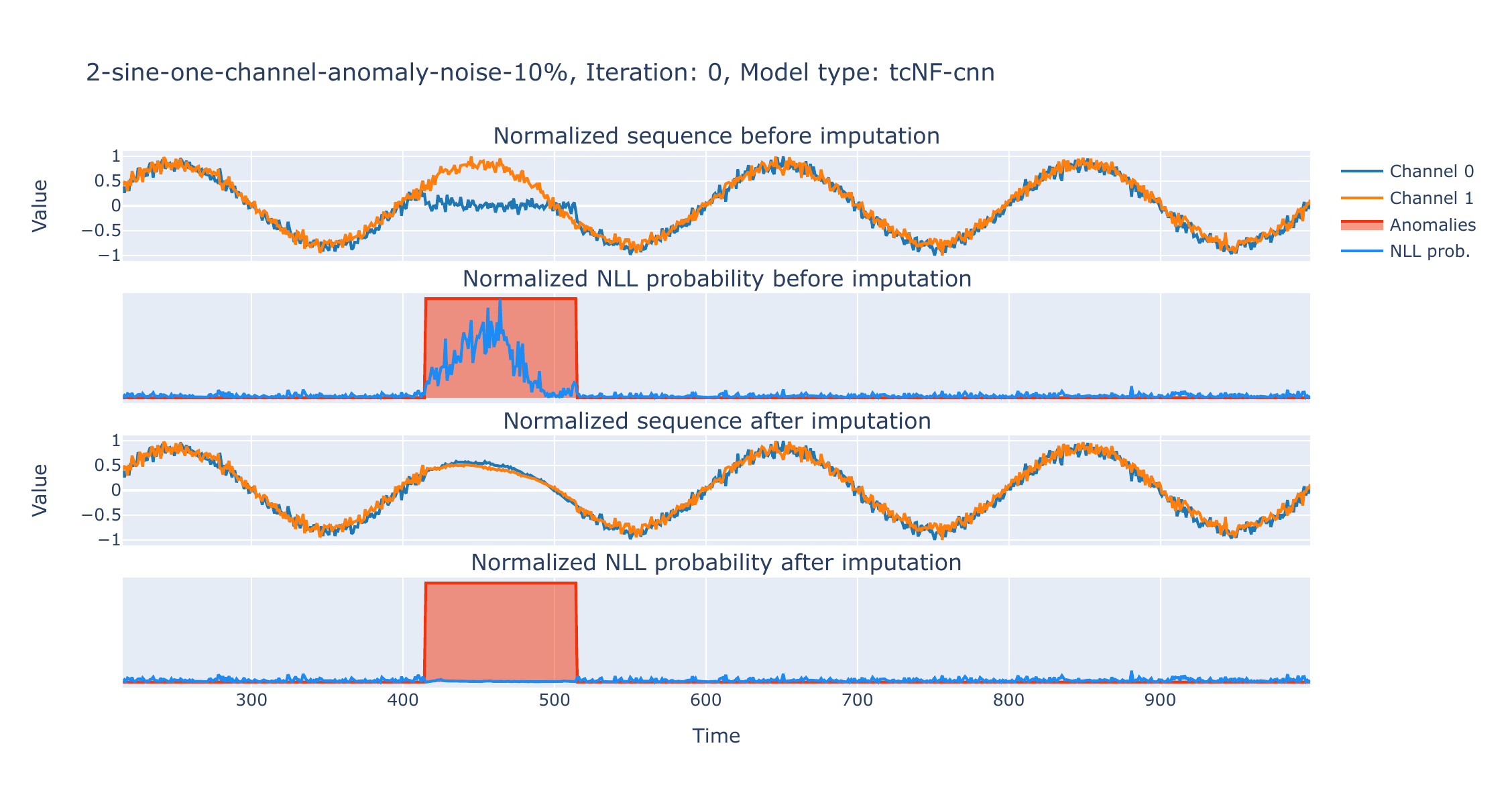}
    \caption{Original and imputed training data with negative log-likelihoods, showing CINDI's long term prediction capabilities and denoising effect.}
\end{figure}

\end{document}